\documentclass[twoside]{article}

\usepackage[accepted]{aistats2023_arxiv}
\usepackage[round]{natbib}

\usepackage{hyperref}
\usepackage{url}
\usepackage{graphicx}
\usepackage{amsmath,amssymb,bm}
\usepackage{color}
\usepackage{subfig}
\usepackage{multirow}
\usepackage{arydshln}
\usepackage[rgb]{KAcolors}

\newcommand{\symbf}[1]{\bm{#1}}

\renewcommand{\arraystretch}{1.15}

\begin{document}

%

%
\runningauthor{Ronny Hug, Stefan Becker, Wolfgang H\"ubner, Michael Arens, J\"urgen Beyerer}

\twocolumn[

\aistatstitle{B\'ezier Curve Gaussian Processes}
\aistatsauthor{ Ronny Hug$^\dagger$ \And Stefan Becker$^\dagger$ \And  Wolfgang H\"ubner$^\dagger$ \And Michael Arens$^\dagger$ \And J\"urgen Beyerer$^{\dagger,\ddagger}$ }
\aistatsaddress{ $^\dagger$ Fraunhofer IOSB and Fraunhofer Center for Machine Learning\\$^\ddagger$ Karlsruhe Institute of Technology (KIT) } ]


\begin{abstract} 
  Probabilistic models for sequential data are the basis for a variety of applications concerned with processing timely ordered information.
  The predominant approach in this domain is given by recurrent neural networks, implementing either an approximate Bayesian approach (e.g. Variational Autoencoders or Generative Adversarial Networks) or a regression-based approach, i.e. variations of Mixture Density networks (MDN).
  In this paper, we focus on the \emph{$\mathcal{N}$-MDN} variant, which parameterizes (mixtures of) probabilistic B\'ezier curves (\emph{$\mathcal{N}$-Curves}) for modeling stochastic processes.
  While in favor in terms of computational cost and stability, MDNs generally fall behind approximate Bayesian approaches in terms of expressiveness.
  Towards this end, we present an approach for closing this gap by enabling full Bayesian inference on top of $\mathcal{N}$-MDNs.
  For this, we show that $\mathcal{N}$-Curves are a special case of Gaussian processes (denoted as $\mathcal{N}$-GP) and then derive corresponding mean and kernel functions for different modalities.
  Following this, we propose the use of the $\mathcal{N}$-MDN as a data-dependent generator for $\mathcal{N}$-GP prior distributions.
  We show the advantages granted by this combined model in an application context, using human trajectory prediction as an example.
\end{abstract}

\section{Introduction}
Models of sequential data play an integral role in a range of different applications related to representation learning, sequence synthesis and prediction.
Thereby, with real-world data often being subject to noise and detection or annotation errors, probabilistic sequence models are favorable.
These take uncertainty in the data into account and provide an implicit or explicit representation of the underlying probability distribution.

The determination of such a probabilistic sequence model is commonly layed out as a learning problem, learning a model of an unknown underlying stochastic process from given sample sequences, which are assumed to be realizations of this process. 
Common approaches are based on either Gaussian Processes \citep{rasmussen2006gaussian} (e.g. \citep{damianou2013deep,mattos2015recurrent}) or more prevalently on neural networks, i.e. approximate Bayesian neural models (e.g. Bayesian Neural Networks \citep{bishop1995neural,blundell2015weight,gal2016dropout}, Variational Autoencoders \citep{kingma2014vae,sohn2015learning,bowman2016generating} and Generative Adversarial Networks \citep{goodfellow2014gan,mirza2014conditional,yu2017seqgan}) or regression-based neural models based on Mixture Density Networks (MDN) \citep{bishop1994mixture} (e.g. \citep{graves2013generating}).
Approximate Bayesian neural models contain stochastic components and allow to directly sample from the modeled stochastic process.
These models typically require computationally expensive Monte Carlo methods during training and inference.
Opposed to that, MDNs are deterministic models, which map a given input onto the parameters of a mixture distribution.
While these models are generally more stable and less computationally heavy during training, Monte Carlo methods are still required for multi-modal inference.
Additionally, due to MDN-based models merely learning to generate point estimates for the target distribution, such models are potentially less expressive as a probabilistic model, e.g. in terms of representing model uncertainty.

In order to tackle difficulties with multi-modal inference in MDN-based probabilistic sequence models, \citet{hug2020introducing} proposed a variation of MDNs, which operate in the domain of parametric curves instead of the data domain, allowing to infer multiple time steps in a single inference step.
The model is built on a probabilistic extension of B\'ezier curves (\emph{$\mathcal{N}$-Curves}), which assume the control points to follow independent Gaussian distributions, thus passing stochasticity to the curve points.
Following this, the \emph{$\mathcal{N}$-MDN} generates a sequence of Gaussian mixture probability distributions in terms of a mixture of $\mathcal{N}$-Curves.

Extending on this approach, in this paper we aim to close the gap between ($\mathcal{N}$-Curve -- based) MDNs and the approximate Bayesian models in terms of expressiveness by establishing a connection between $\mathcal{N}$-MDNs and the Gaussian process (GP) framework.
Our basic idea revolves around employing the $\mathcal{N}$-MDN for determining a data-dependent GP prior based on $\mathcal{N}$-Curves.
To achieve this, we first show that the underlying $\mathcal{N}$-Curves are a special case of Gaussian processes.
We denote this $\mathcal{N}$-Curve -- induced GP as $\mathcal{N}$-GP.
Following this, we derive mean and kernel functions for the $\mathcal{N}$-GP considering different modalities, i.e. univariate, multivariate and multi-modal Gaussian processes.
Ultimately, this allows for a more expressive and flexible probabilistic model by employing the GP framework, with the benefits of a regression-based model. 

In our evaluation, we explore the advantages granted by our combined model from a practical perspective.
Following this, using human trajectory prediction as an exemplary sequence predicton task, we use our model for manipulating predictions generated by the underlying $\mathcal{N}$-MDN by calculating different posterior distributions according to the induced $\mathcal{N}$-GP.
For the posterior distributions, we consider two use-cases.
First, improving the overall prediction performance by calculating the predictive posterior distribution given one or more observed trajectory points.
Second, we explore the possibilities of updating the predictions under the presence of new measurements within the prediction time horizon.
In our approach, this does not require any additional passes through the $\mathcal{N}$-MDN.

To summarize, our main contributions are given by:
\begin{enumerate}
	\item A proof for probabilistic B\'ezier curves ($\mathcal{N}$-Curves) being Gaussian processes.
	\item The derivation of GP mean and covariance functions induced by $\mathcal{N}$-Curves, covering the univariate, multivariate and multi-modal cases.
	\item A probabilistic sequence model, which combines the stability and low computational complexity of $\mathcal{N}$-Curve -- based MDNs with the expressiveness and flexibility of Gaussian processes.
\end{enumerate}

\section{Preliminaries}

\subsection{Gaussian Processes} 
A Gaussian process (GP, \citep{rasmussen2006gaussian}) is a stochastic process $\{X_t\}_{t \in T}$ with index set $T$, where the joint distribution of stochastic variables $X_{t_i}$ for an arbitrary, finite subset $\{t_1, ..., t_N\}$ of $T$ is a multivariate Gaussian distribution. 
For simplicity, we will interpret the index as \emph{time} throughout this paper.
The joint distribution is obtained using an explicit mean function $m(t)$ and positive definite covariance function $k(t_i,t_j) = \text{cov}(f(t_i),f(t_j))$, commonly referred to as the kernel of the Gaussian process, and yields a multivariate Gaussian prior probability distribution over function space.
Commonly, $m(t) = 0$ is assumed.
Given a collection of sample points $X_*$ of a function $f(t)$, the posterior (predictive) distribution $p(X|X_*)$ modeling non-observed function values $X$ can be obtained.
As such, Gaussian processes provide a well-established model for probabilistic sequence modeling.

\subsection{Probabilistic B\'ezier Curves}
\label{ss:n-curve} 
Probabilistic B\'ezier Curves ($\mathcal{N}$-Curves, \citep{hug2020introducing}) are B\'ezier curves \citep{prautzsch2002bezier} defined by $(L + 1)$ independent $d$-dimensional Gaussian control points $\mathcal{P} = \{P_0, ..., P_L\}$ with $P_l \sim \mathcal{N}(\symbf{\mu}_l,\symbf{\Sigma}_l)$.
Through the curve construction function
\begin{align}
	\label{eq:BN}
		X_t = B_{\mathcal{N}}(t, \mathcal{P}) = (\mu_\mathcal{P}(t), \Sigma_\mathcal{P}(t))
	\end{align}
	with
	\begin{align}
	\label{eq:mu}
		\mu_\mathcal{P}(t) = \sum_{l=0}^{L} b_{l,L}(t) \symbf{\mu}_l
	\end{align}
	and
	\begin{align}
	\label{eq:psi}
		\Sigma_\mathcal{P}(t) &= \sum_{l=0}^{L} \left(b_{l,L}(t)\right)^2 \symbf{\Sigma}_l,
	\end{align}
where 
\begin{align}
	b_{l,L}(t) = \binom{L}{l} (1-t)^{L-l}t^l
\end{align}
are the Bernstein polynomials \citep{lorentz2013bernstein}, the stochasticity is passed from the control points to the curve points $X_t \sim \mathcal{N}(\mu_\mathcal{P}(t),\Sigma_\mathcal{P}(t))$, yielding a sequence of Gaussian distributions $\{X_t\}_{t \in [0,1]}$ along the underlying B\'ezier curve.
Thus, a stochastic process with index set $T = [0,1]$ can be defined.
For representing discrete data, i.e. sequences of length $N$, a discrete subset of $T$ can be employed for connecting sequence indices with evenly distributed values in $[0,1]$, yielding 
\begin{align} 
\label{eq:T}
	T_N = \left\{\frac{v}{N-1}|v \in \left\{0, ..., N-1\right\}\right\} = \left\{t_1, ..., t_N\right\}.
\end{align}

\section{\texorpdfstring{$\mathcal{N}$-}{Probabilistic B\'ezier }Curve Gaussian Processes}
With $\mathcal{N}$-Curves providing a representation for stochastic processes $\{X_t\}_{t \in T}$ comprised of Gaussian random variables $X_t \sim \mathcal{N}(\symbf{\mu}, \symbf{\Sigma})$, we first show that $\mathcal{N}$-Curves are a special case of GPs with an implicit covariance function.
Following the definition of GPs \citep{mackay2003information,rasmussen2006gaussian}, an $\mathcal{N}$-Curve can be classified as a GP, if for any finite subset $\{t_1, ..., t_N\}$ of $T$, the joint probability density $p(X_{t_1}, ..., X_{t_N})$ of corresponding random variables is Gaussian.
This property is referred to as the \emph{GP property}.
We show that this property holds true by reformulating the curve construction formula into a linear transformation\footnote{For clarity, multivariate random variables may be written in bold font occasionally.} $\bm{X} = \mathbf{C} \cdot \bm{P}$ of the Gaussian control points stacked into a $((L+1) \cdot d \times 1)$ vector  
\begin{align}
	\bm{P}^\top = 
	\begin{pmatrix}
		P^\top_0 &
		P^\top_1 &
		\cdots &
		P^\top_{L}
	\end{pmatrix}
\end{align}
using a $(N \cdot d \times (L+1) \cdot d)$ transformation matrix 
\begin{align}
	\mathbf{C} = 
	\begin{pmatrix}
		\mathbf{B}_{0,L}(t_1) & \dots & \mathbf{B}_{L,L}(t_1) \\
		\vdots & \ddots & \vdots \\
		\mathbf{B}_{0,L}(t_N) & \dots & \mathbf{B}_{L,L}(t_N)
	\end{pmatrix}
\end{align}
determined by the Bernstein polynomials, with $\mathbf{B}_{l,L}(t_j) = b_{l,L}(t_j) \mathbf{I}_d$ and $t_j \in T_N$.
As $\bm{P}$ is itself a Gaussian random vector, $\bm{X}$ is again Gaussian with its corresponding probability density function $p(\bm{X}) = p(X_1, ..., X_N)$ being a Gaussian probability density.

As the Gaussians along an $\mathcal{N}$-Curve are correlated through the use of common control points, the mean and kernel functions of the induced GP, denoted as $\mathcal{N}$-GP in the following, can be given explicitly.
In the following sections, we thus derive the $\mathcal{N}$-GP for the univariate, multivariate and multi-modal case, with respective mean and kernel functions.
Afterwards, we discuss practical implications of the derived Gaussian process variant.

\subsection{Univariate \texorpdfstring{$\mathcal{N}$-}{Probabilistic B\'ezier }Curve Gaussian Processes}
Being the most common use case, we first consider univariate GPs, which target scalar-valued functions $f: \mathbb{R} \rightarrow \mathbb{R}$.
Besides that, it grants a simple case for deriving the mean and kernel functions induced by a given $\mathcal{N}$-Curve while also allowing a visual examination of some properties of the $\mathcal{N}$-GP.
Here, the stochastic control points $P_i$ are defined by the mean value $\mu_i$ and variance $\sigma^2_i$.
The mean function is equivalent to Eq. \ref{eq:mu}.
Thus, we focus on the kernel $k_{\mathcal{P}}(t_i,t_j)$ for two curve points $X = f(t_i) = \sum^{L}_{l=0} b_{l,L}(t_i) P_l$ and $Y = f(t_j) = \sum^{L}_{l=0} b_{l,L}(t_j) P_l$ at indices $t_i$ and $t_j$ with $t_i,t_j \in [0,1]$.
The respective mean values are given by $\mu_X = \sum^{L}_{l=0} b_{l,L}(t_i) \mu_l$ and $\mu_Y = \sum^{L}_{l=0} b_{l,L}(t_j) \mu_l$.
From $k(t_i,t_j) = \text{cov}(f(t_i),f(t_j))$ then follows:
\begin{align*}
	\begin{split}
		k&_{\mathcal{P}}(t_i,t_j) = \mathbb{E}[(X - \mu_X)(Y - \mu_Y)] \\
		&= \mathbb{E}\left[\left(\sum^{L}_{l=0} b_{l,L}(t_i) P_l - \mu_X\right)\left(\sum^{L}_{l=0} b_{l,L}(t_j) P_l - \mu_Y\right)\right] \\
		&= \mathbb{E}\left[\sum^{L}_{l=0} b_{l,L}(t_i) b_{l,L}(t_j) P^2_l \right] \\
    &+ \mathbb{E} \left[ \sum^{L}_{l=0} \left( \sum^{L}_{l'=0,l' \neq l} b_{l,L}(t_i) b_{l',L}(t_j) P_l P_{l'} \right)\right] \\
		&- \underbrace{\mu_Y\sum^{L}_{l=0} b_{l,L}(t_i)\mu_l}_{=\mu_Y\mu_X} - \underbrace{\mu_X\sum^{L}_{l=0} b_{l,L}(t_j)\mu_l}_{=\mu_X\mu_Y} + \mu_X\mu_Y. 
	\end{split}
\end{align*}

With $\mathbb{E}[P_i \cdot P_j] = \mathbb{E}[P_i] \cdot \mathbb{E}[P_j]$, which follows from the independence of the control points, and $\mathbb{E}[P^2_i] = \text{Var}[P_i] + (\mathbb{E}[P_i])^2$, follows the closed-form solution
\begin{align}
	\label{eq:univ_kernel}
\begin{split}
	k&_{\mathcal{P}} (t_i,t_j) 
	= \sum^{L}_{l=0} b_{l,L}(t_i) b_{l,L}(t_j) (\sigma^2_l + \mu^2_l) \\
	&+ \sum^{L}_{l=0} \left( \sum^{L}_{l'=0,l' \neq l} b_{l,L}(t_i) b_{l',L}(t_j) \mu_l \mu_{l'} \right) - \mu_X\mu_Y. 
\end{split}
\end{align}

\begin{figure*}[tb]
	\centering
	\includegraphics[width=0.22\textwidth]{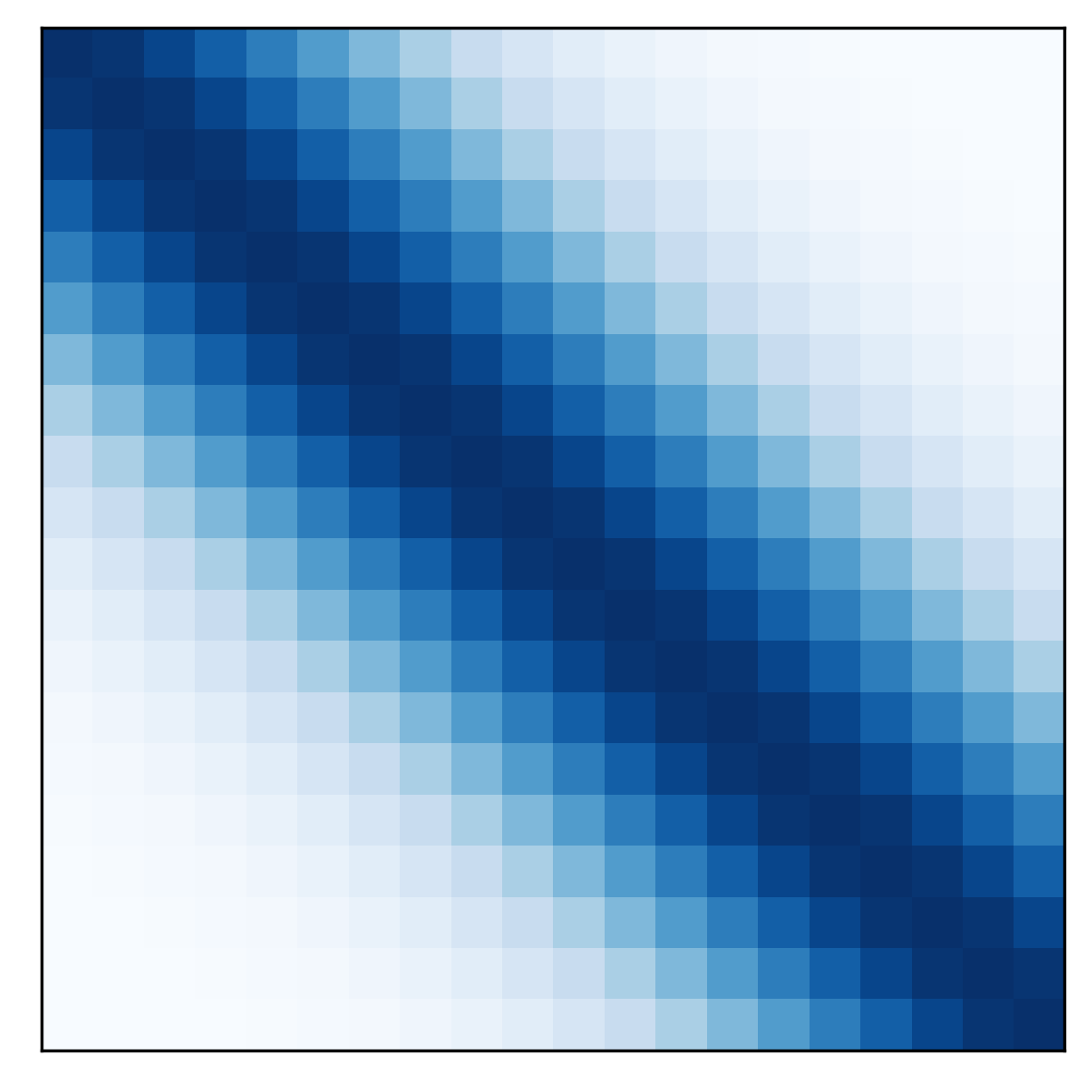} 
	\hspace{9pt}
	\includegraphics[width=0.22\textwidth]{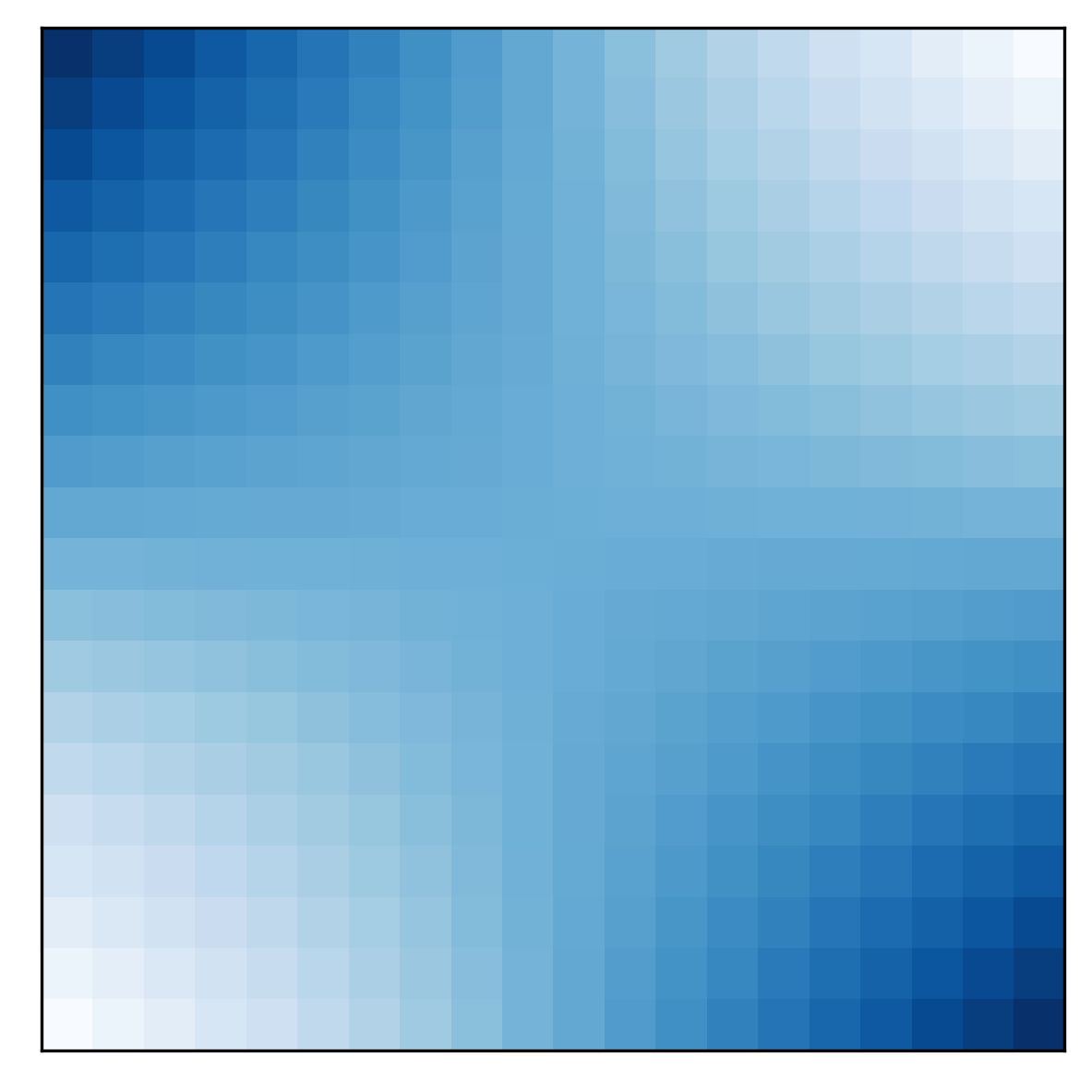}
	\hspace{9pt}
	\includegraphics[width=0.22\textwidth]{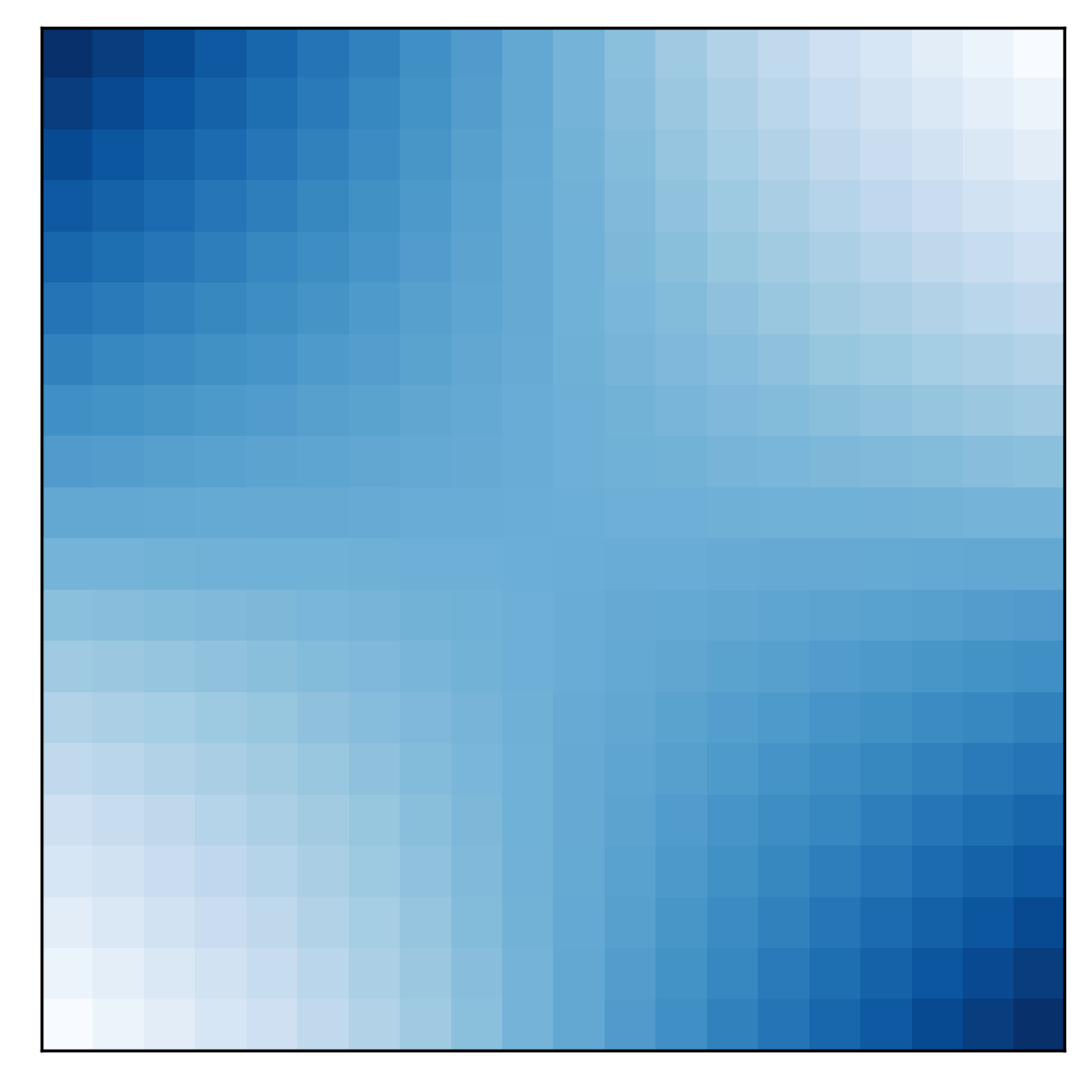}
	\hspace{9pt}
	\includegraphics[width=0.22\textwidth]{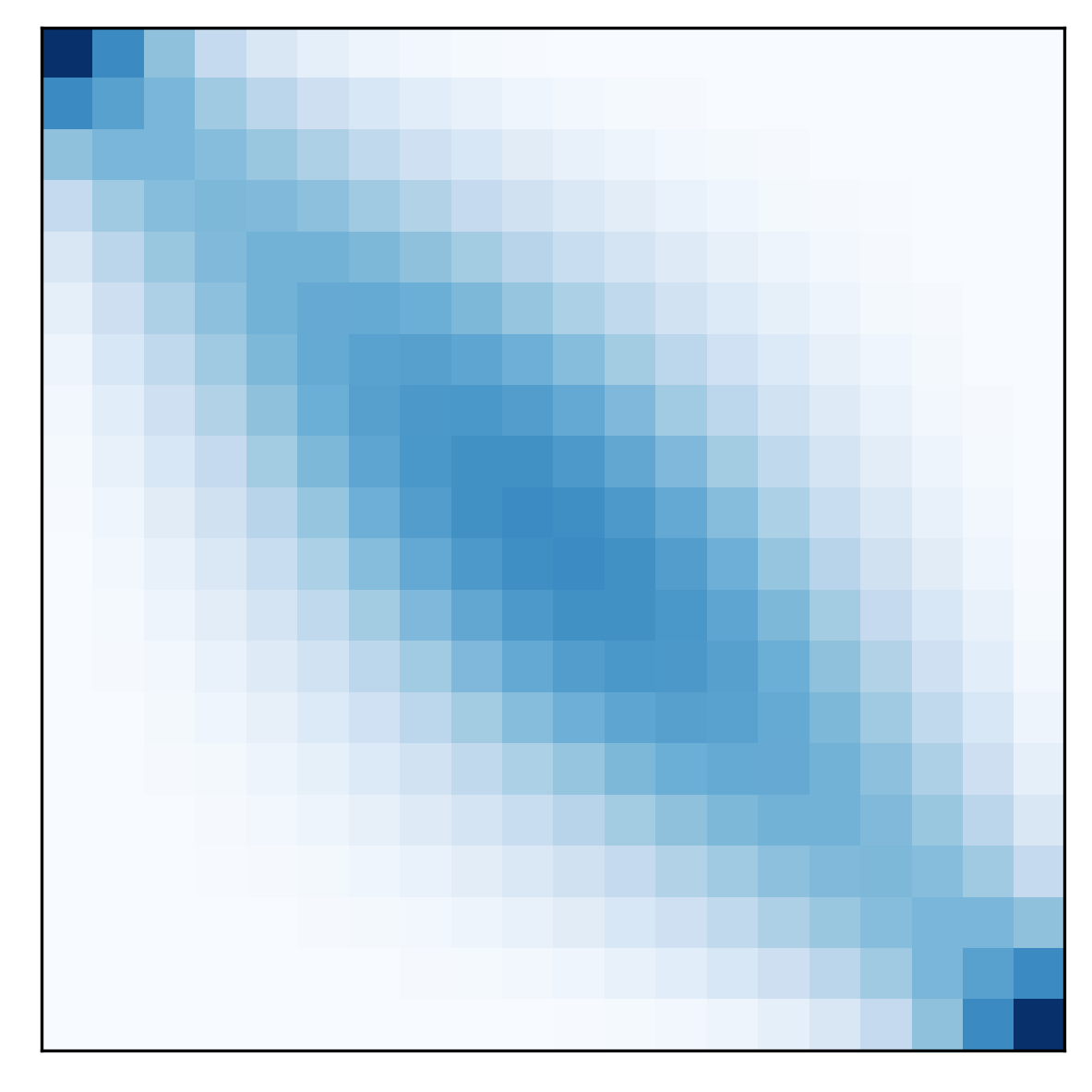}
	\caption{Gram matrices for $20$ equally spaced values in $[0,1]$ obtained by using different GP kernels.
	     	Left to right: RBF kernel, linear kernel and $\mathcal{N}$-GP kernels $k_{\mathcal{P}_1}(t_i,t_j)$ and $k_{\mathcal{P}_2}(t_i,t_j)$.}
	\label{fig:kernels}
\end{figure*}

As the $\mathcal{N}$-GP is heavily dependent on the given set of control points, it allows for a range of different kernels.
For comparison, Fig. \ref{fig:kernels} illustrates two standard kernels \citep{gortler2019visual}, given by a \emph{radial basis function} (RBF) kernel 
\begin{align}
	k^{\text{rbf}}_{\sigma,l}(t_i,t_j) = \sigma^2 \exp \left( -\frac{\|t_i-t_j\|^2}{2l^2} \right),
\end{align}
with $\sigma=1$ and $l=0.25$ and a linear kernel
\begin{align}
	k^{\text{lin}}_{\sigma,\sigma_b,c}(t_i,t_j) = \sigma^2_b + \sigma^2(t_i - c)(t_j - c),
\end{align} 
with $\sigma = \sigma_b = c = 0.5$, and two $\mathcal{N}$-GP kernels $k_{\mathcal{P}_1}(t_i,t_j)$ and $k_{\mathcal{P}_2}(t_i,t_j)$.
$\mathcal{P}_1$ consists of two unit Gaussians and $\mathcal{P}_2$ consists of $9$ zero mean Gaussian control points with standard deviations $\sigma_0 = \sigma_8 = 1$, $\sigma_1 = \sigma_7 = 1.25$, $\sigma_2 = \sigma_6 = 1.5$, $\sigma_3 = \sigma_5 = 1.75$ and $\sigma_4 = 2$.
The standard deviations vary in order to cope with non-linear blending (see Eq. \ref{eq:psi}).
The Gram matrices calculated from $20$ equally spaced values in $[0, 1]$ are depicted for each kernel.

\begin{figure*}[tb]
	\centering
	\includegraphics[width=0.24\textwidth] {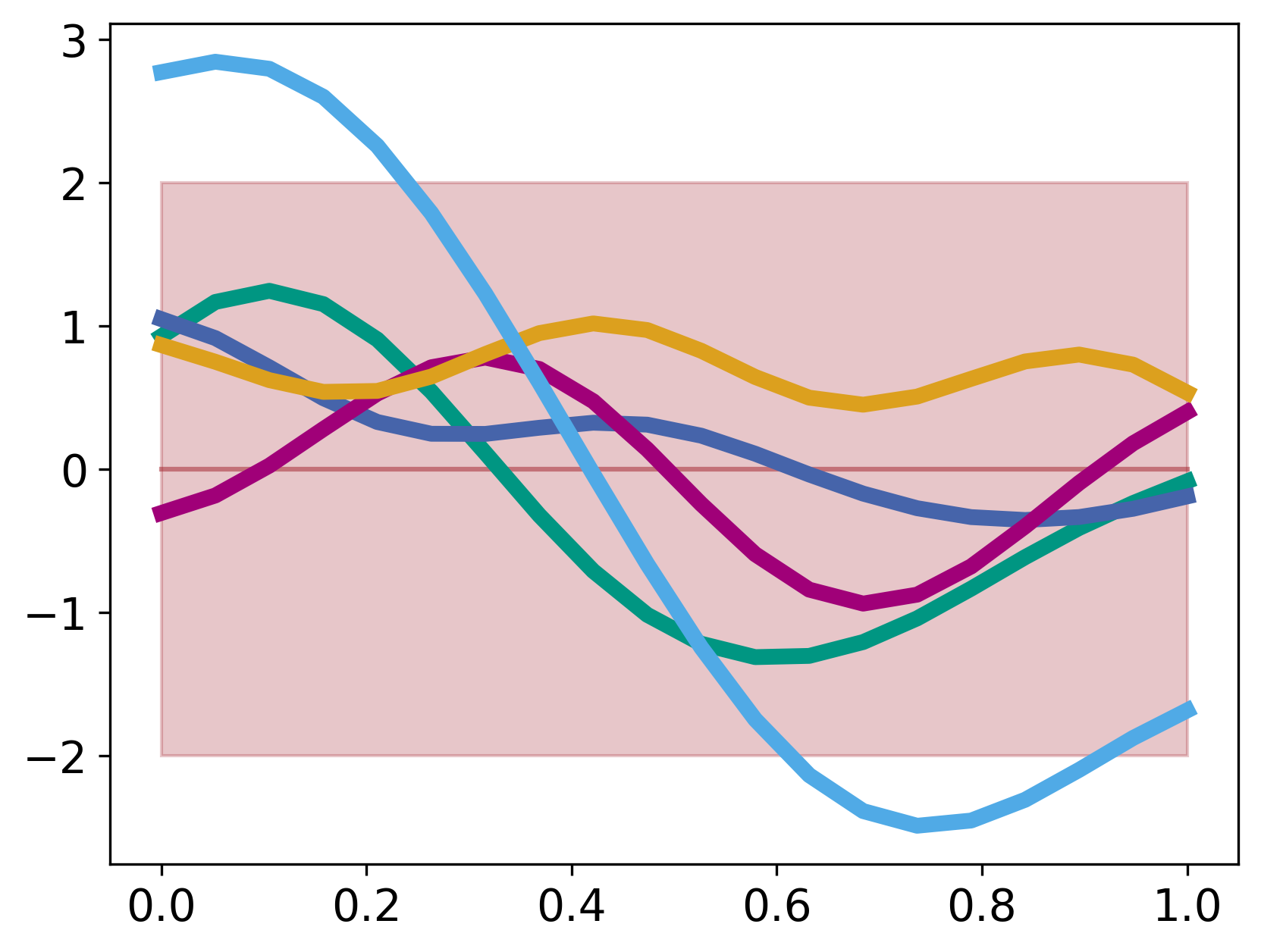} 
	\includegraphics[width=0.24\textwidth] {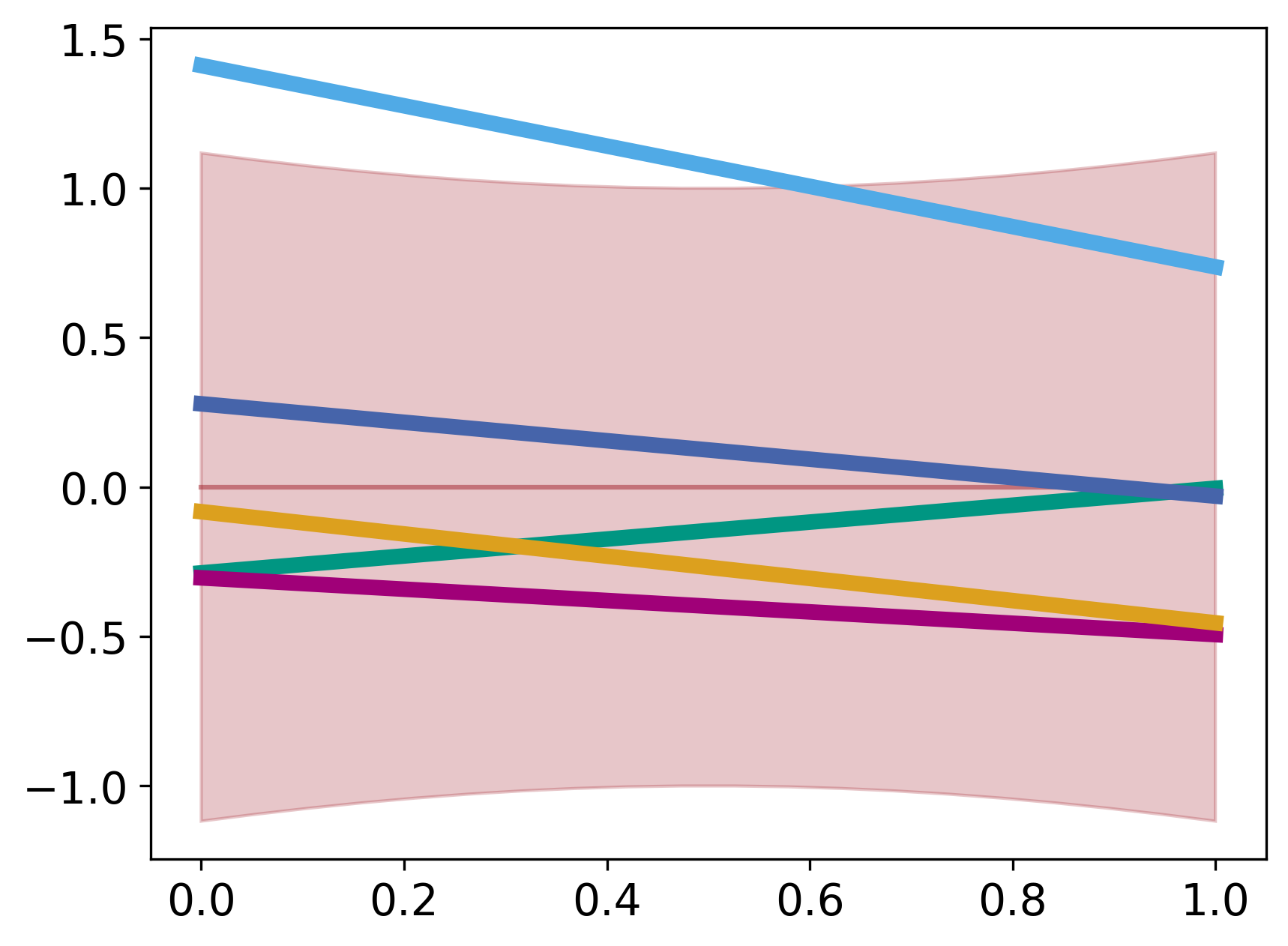}
	\includegraphics[width=0.24\textwidth] {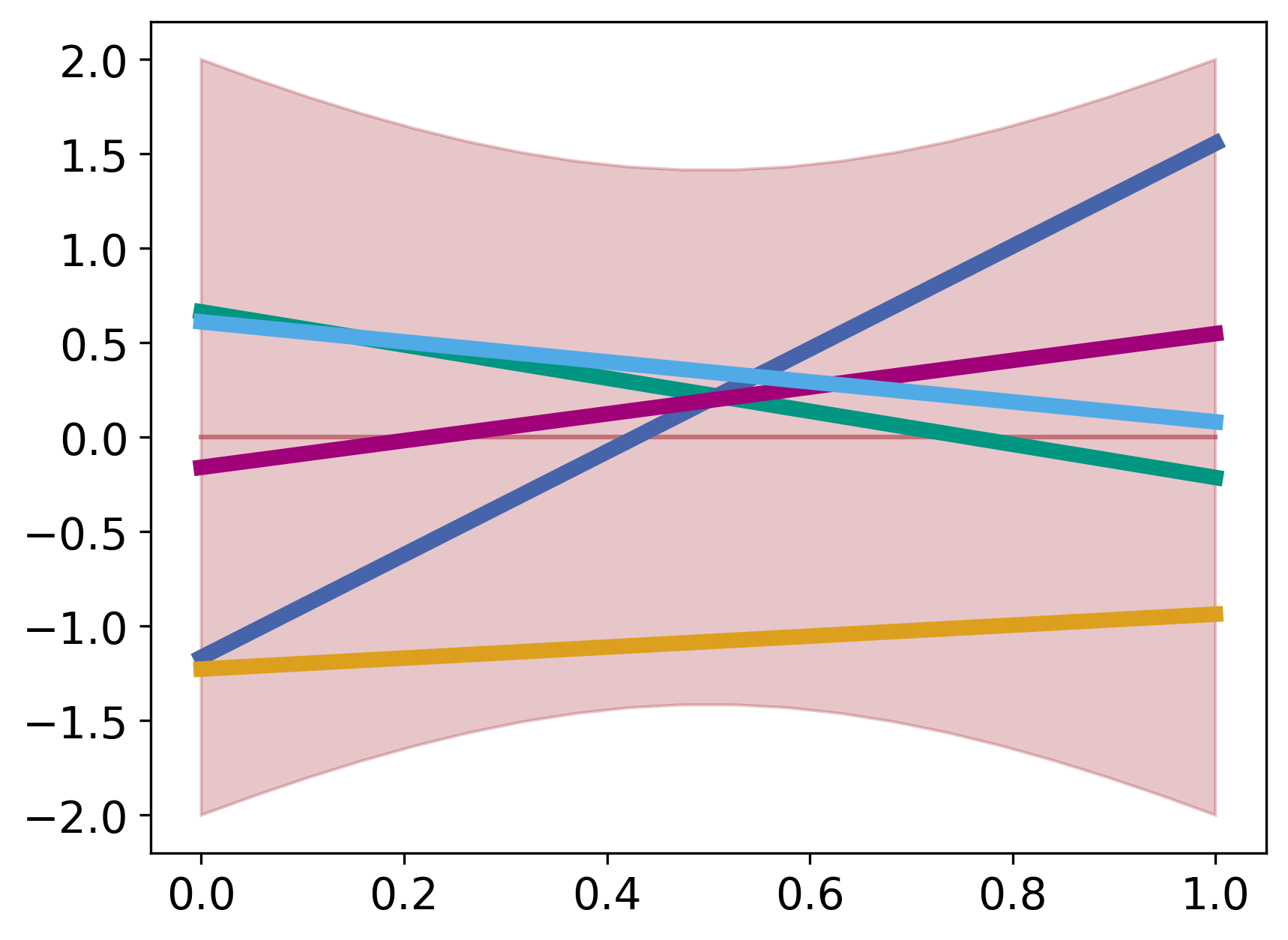}
	\includegraphics[width=0.24\textwidth] {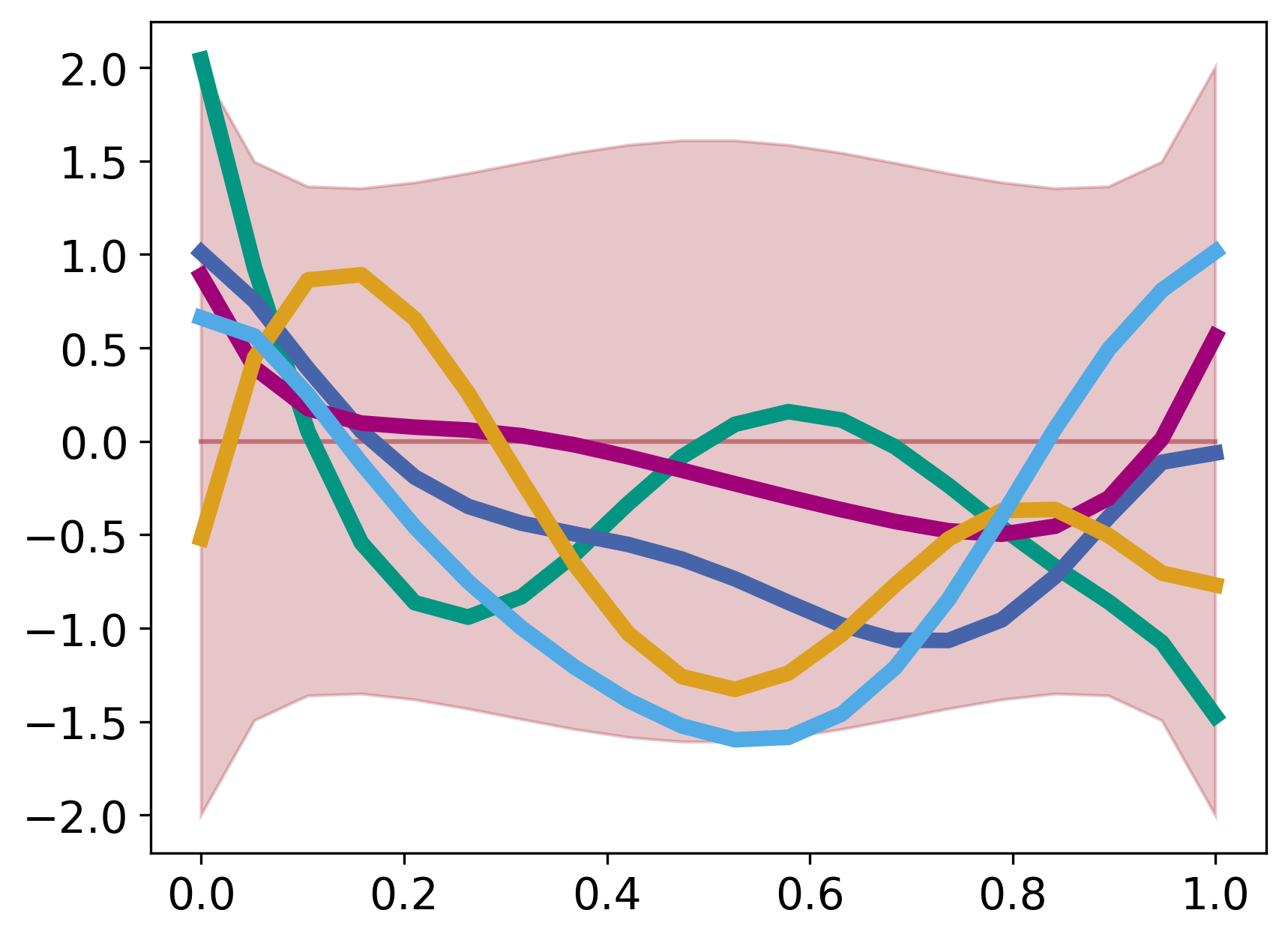}
	\caption{Samples drawn from prior distributions using different GP kernels.
	         The $2\sigma$ region is depicted as a red shaded area.
			 Left to right: RBF kernel, linear kernel and $\mathcal{N}$-GP kernels $k_{\mathcal{P}_1}(t_i,t_j)$ and $k_{\mathcal{P}_2}(t_i,t_j)$.}
	\label{fig:prior_samples}
\end{figure*}

When comparing the Gram matrices, it can be seen, that the matrix calculated with $k_{\mathcal{P}_1}$ is equal to that calculated with $k^{\text{lin}}_{\sigma=\sigma_b=c=0.5}(t_i,t_j)$ when normalizing its values to $[0, 1]$.
On the other hand, the matrix obtained with $k_{\mathcal{P}_2}$, which is derived from a more complex $\mathcal{N}$-Curve, tends to be more comparable to the matrix calculated with $k^{\text{rbf}}_{\sigma=1,l=0.25}(t_i,t_j)$.
These parallels are also visible when comparing sample functions drawn from each GP prior, assuming a zero mean GP using the different kernels as depicted in Fig. \ref{fig:prior_samples}.


\subsection{Multivariate \texorpdfstring{$\mathcal{N}$-}{Probabilistic B\'ezier }Curve Gaussian Processes} 
Multivariate GPs target vector-valued functions $\mathbf{f}(t)$, which map scalar inputs onto $d$-dimensional vectors, e.g. $\mathbf{f}: \mathbb{R} \rightarrow \mathbb{R}^d$.
Following this, for elevating our univariate $\mathcal{N}$-GP derived in the previous section to the multivariate case, there exist two closely related approaches we can adopt.
The first sticks with the multivariate Gaussian distribution and models matrix-valued random variables by using stacked mean vectors in combination with block partitioned covariance matrices \citep{alvarez2012kernels}.
The other revolves around the matrix normal distribution \citep{chen2020remarks,chen2020multivariate}, which can be transformed into a multivariate Gaussian distribution by vectorizing the mean matrix and calculating the covariance matrix as the Kronecker product of both scale matrices, thus establishing a connection to the former.

We adopt the first approach, as it simplifies the extension of the univariate $\mathcal{N}$-GP.
Following this, the Gram matrix of a $d$-variate GP for a finite index subset with $|T_N| = N$ is given by the $(Nd \times Nd)$ block partitioned matrix 
\begin{align}
	\Sigma = \begin{pmatrix}
		\mathbf{K}_\mathcal{P}(t_1,t_1) & \cdots & \mathbf{K}_\mathcal{P}(t_1,t_N) \\
		\vdots & \ddots & \vdots \\
		\mathbf{K}_\mathcal{P}(t_N,t_1) & \cdots & \mathbf{K}_\mathcal{P}(t_N,t_N)
	\end{pmatrix}
\end{align}
calculated using the matrix-valued kernel $\mathbf{K}_\mathcal{P}(t_i,t_j) = \text{cov}(\bm{X}, \bm{Y})$.
Here, $\bm{X} = \sum^{L}_{l=0} b_{l,L}(t_i) \bm{P}_l$ and $\bm{Y} = \sum^{L}_{l=0} b_{l,L}(t_j) \bm{P}_l$ are now $d$-variate Gaussian random variables resulting from the linear combination of $d$-variate $\mathcal{N}$-Curve control points $\bm{P}_l$.
Thus, the multivariate generalization of Eq. \ref{eq:univ_kernel} yields a $(d \times d)$ matrix and is given by
\begin{align}
	\label{eq:ngp_cov} 
\begin{split}
	\mathbf{K}_\mathcal{P}&(t_i,t_j) = \mathbb{E}[(\bm{X} - \symbf{\mu}_X)(\bm{Y} - \symbf{\mu}_Y)^\top] \\
	&= \mathbb{E}\left[ \bm{X}\mathbf{Y}^\top \right] - \underbrace{\mathbb{E}\left[ \bm{X}\symbf{\mu}^\top_Y \right]}_{=\symbf{\mu}_X\symbf{\mu}^\top_Y} - \underbrace{\mathbb{E}\left[ \symbf{\mu}_X\bm{Y}^\top \right]}_{=\symbf{\mu}_X\symbf{\mu}^\top_Y} + \symbf{\mu}_X\symbf{\mu}^\top_Y\\
	&= -\symbf{\mu}_X\symbf{\mu}^\top_Y + \sum^{L}_{l=0} b_{l,L}(t_i) b_{l,L}(t_j) \left( \symbf{\Sigma}_l + \symbf{\mu}_l \symbf{\mu}^\top_l \right) \\
	&+ \sum^{L}_{l=0} \left( \sum^{L}_{l'=0,l' \neq l} b_{l,L}(t_i) b_{l',L}(t_j) \symbf{\mu}_l \symbf{\mu}^\top_{l'} \right).
\end{split}
\end{align}

The $(n \times d)$ mean vector is defined as the concatenation of all point mean vectors $\symbf{\mu}_\mathcal{P}(t_i)$ (see also Eq. \ref{eq:mu}), i.e.
\begin{align}
	\label{eq:ngp_mean}
	\mathbf{m}_\mathcal{P}(T_N) = \begin{pmatrix}
		\symbf{\mu}_\mathcal{P}(t_1) \\
		\vdots \\
		\symbf{\mu}_\mathcal{P}(t_N)
	\end{pmatrix}.
\end{align}

\subsection{Multi-modal \texorpdfstring{$\mathcal{N}$-}{Probabilistic B\'ezier }Curve Gaussian Processes} 
With sequence modeling tasks often being multi-modal problems and GPs as presented before being incapable of modeling such data, we consider multi-modal GPs as a final case.
A common approach to increasing the expressiveness of a statistical model, e.g. for heteroscedasticity or multi-modality, is given by mixture modeling approaches.
Thereby, rather than a single model or distribution, a mixture of which are used with each component in the mixture covering a subset of the data.
Generaly speaking, a widely used mixture model is given by the Gaussian mixture model \citep{bishop2006pattern}, which is defined as a convex combination of $K$ Gaussian distributions with (mixing) weights $\symbf{\pi} = \{\pi_1, ..., \pi_K\}$ and probability density function
\begin{align}
\begin{split}
	p(\mathbf{x}) = \sum^K_{k=1} \underbrace{p(z=k)}_{\pi_k}\underbrace{p(\mathbf{x}|z=k)}_{\mathcal{N}(\symbf{\mu}_k, \symbf{\Sigma}_k)},~z \sim \text{Categorical}(\symbf{\pi}).
\end{split}
\end{align}
In the case of GPs, a popular approach is given by the \emph{mixture of Gaussian process experts} \citep{tresp2000mixtures,rasmussen2001infinite,yuan2008variational}, which extends on the mixture of experts model \citep{jacobs1991adaptive}.
In this approach, the mixture model is comprised of a mixture of $K$ GP \emph{experts} (components) $\mathcal{G}_k$ with mean function $\mathbf{m}_k$ and kernel $\mathbf{K}_k$
\begin{align}
	\sum^K_{k=1} p(z=k|\mathbf{x})\mathcal{G}_k(\mathbf{m}_k(\cdot),\mathbf{K}_k(\cdot,\cdot)),
\end{align}
weighted using a conditional weight distribution $\text{Categorical}(\symbf{\pi}|\mathbf{x})$ for a given sample $\mathbf{x}$.
The weight distribution is generated by a \emph{gating network}, which decides on the influence of each local \emph{expert} for modeling a given sample.
This is the key difference to the Gaussian mixture model, where the weight distribution is static and determined a priori (e.g. via EM \citep{dempster1977maximum} or an MDN \citep{bishop1994mixture}).
It can be noted that mixtures of experts are also often used to lower the computational load of a GP model, as less data points have to be considered during inference due to the use of local experts (e.g. \citep{deisenroth2015distributed,lederer2021gaussian}).

In line with the approach given in \citep{hug2020introducing}, which builds on Gaussian mixture models, we define the multi-modal extension of our $\mathcal{N}$-GP as a mixture of $K$ $\mathcal{N}$-GPs 
\begin{align}
	\label{eq:ngp_mix}
	\mathcal{M}\mathcal{G}(\symbf{\pi},\{\mathcal{G}_k\}_{k \in \{1,...,K\}}) = \sum^K_{k=1}\pi_k\mathcal{G}_k(\symbf{\mu}_{\mathcal{P}_k}(\cdot), \mathbf{K}_{\mathcal{P}_k}(\cdot,\cdot)),
\end{align}
with $\mathcal{N}$-GP components $\mathcal{G}_k$ and the prior weight distribution $\symbf{\pi} = \{\pi_1,...\pi_K\}$ with $\sum^K_{k=1}\pi_k = 1$.
Here, the mean and kernel functions are determined separately for each GP component according to equations \ref{eq:mu} and \ref{eq:univ_kernel} in the unimodal case, or \ref{eq:ngp_mean} and \ref{eq:ngp_cov} in the multi-modal case.
Given these functions and the weights $\pi$, the mixture distribution can be evaluated at a given index.

\subsection{Practical Implications} 
\label{sec:implications}
In order to apply $\mathcal{N}$-GPs to sequence modeling tasks, we propose to combine $\mathcal{N}$-GPs with $\mathcal{N}$-MDNs.
In this combined model, we treat the $\mathcal{N}$-MDN as a data-dependent generator for prior distributions within the GP framework. 
As a result, full Bayesian inference is enabled in the otherwise regression-based neural model, combining the stability and low computational complexity of $\mathcal{N}$-MDNs with the expressiveness and flexibility of Gaussian processes.

To put the value of our proposed model into perspective from a practical standpoint, we discuss different use-cases we expect to benefit from the $\mathcal{N}$-GP extension.
For conciseness, we focus on the task of sequence prediction, where given a length $N_\text{in}$ input sequence (the \emph{observation}), the subsequent $N_\text{pred}$ sequence elements need to be predicted.
The length of each sequence is referred to as the observation and prediction time horizon, respectively.
As a technical detail, in our combined model, the $\mathcal{N}$-MDN is tasked to model the input sequence in addition to the actual sequence to be predicted.
In this way, we ensure that within the GP framework, we are able to condition on elements within the input sequence.
Finally, we are considering two use-cases, i.e. prediction \emph{refinement} and \emph{update}.

\paragraph{Refinement}
We first examine the $\mathcal{N}$-GP as a tool for improving the overall prediction performance considering different posterior predictive distributions given different subsets of the input sequence. 
We expect this refinement to improve the prediction performance, as the original maximum likelihood prediction generated by the $\mathcal{N}$-MDN tends to average out small variations in the data and the refinement procedure adapts the prediction more towards the actual observation.
We consider this the most practically relevant use-case, as it directly affects the model's accuracy.

\paragraph{Update}
Another interesting option opened up by embedding the $\mathcal{N}$-MDN into the GP framework is given by the use-case of updating a multi-step prediction under the presence of new data within the prediction time horizon.
As sequence models usually predict several time steps into the future, an easy to calculate and fast to compute update to the prediction under the presence of new data can be valuable.
The $\mathcal{N}$-GP enables such updates without the need for additional passes through the underlying neural network.
Further, it is unaffected by potentially missing intermediate observations. 
This is especially valuable, as common sequence prediction models require complete sequences as input.
Thus, such gaps need to be filled with information extracted from the model's own initial prediction, which can be problematic under the presence of multiple modes in the predicted distribution, making Monte Carlo methods a necessity. 
In contrast, within the GP framework missing information between observed data points is naturally interpolated.
It is worth noting, that when certain requirements are met, our model allows to fill gaps in light of fragmented observations easily.
First, a full $N_\text{in}$-step input sequence is required for the underlying $\mathcal{N}$-MDN for generating our prior distribution.
Second, the gaps to fill must be within the modeled prediction time horizon.

\section{Evaluation}  
Throughout this section we aim to support our proposed combination of $\mathcal{N}$-MDNs\footnote{We omit a state-of-the-art comparison of the $\mathcal{N}$-MDN, as it has been proven viable and competetive on the given task \citep{hug2020introducing} (see also supplemental material).} and $\mathcal{N}$-GPs, considering the \emph{refinement} and \emph{update} use-cases in the scope of an established sequence processing task, i.e. human trajectory prediction, using standard benchmarking datasets.
This task provides easy to interpret and visualize results while also providing a lot of complexity being a highly multi-modal problem, despite its low data dimensionality.
In human trajectory prediction, given $N_\text{in}$ points of a trajectory as input, a sequence model is tasked to predict the subsequent $N_\text{pred}$ trajectory points.
As indicated before, the $\mathcal{N}$-MDN models both, the observed and to be predicted trajectory, in order to enable GP-based inference.

\subsection{Parameter Estimation and Conditional Inference} 
Our $\mathcal{N}$-GP relies on prior distributions generated by an $\mathcal{N}$-MDN, whose parameters are learned from data. 
Following \citet{hug2020introducing}, the $\mathcal{N}$-MDN is a feedfoward neural network, which maps an input vector $\mathbf{v}$ onto the parameters of a $K$-component $\mathcal{N}$-Curve mixture, i.e. the weights and Gaussian curve control points.
Here, the vector $\mathbf{v}$ is a representation of the input trajectory obtained by applying a sequence encoder.
In line with the original approach, an LSTM \citep{hochreiter1997long} is applied, which is a common choice in human trajectory prediction.
For training the LSTM - MDN combination using a set of $M$ fixed-length trajectories $\mathcal{D} = \{ \mathcal{X}_1, ..., \mathcal{X}_M \}$ with $\mathcal{X}_i = \{ \mathbf{x}^i_1, ..., \mathbf{x}^i_N \}$, the negative log-likelihood loss
\begin{align}
\begin{split}
	\mathcal{L} &= \frac{1}{M} \sum_{j=1}^{M} -\log~\sum_{k=1}^{K} \exp \Big\{ \log \pi_k + \\
	& \sum_{i=1}^{N} \log \left( \mathcal{N}(\mathbf{x}^{j}_i|\mu_{\mathcal{P}}(t_i),\Sigma_{\mathcal{P}}(t_i)) \right)\Big\}
\end{split}
\end{align}
is applied in conjunction with a gradient descent policy.

Then, given an input trajectory, the $\mathcal{N}$-MDN generates an $\mathcal{N}$-Curve mixture, which models the input as well as possible future trajectories.
Using Equations \ref{eq:ngp_cov}, \ref{eq:ngp_mean} and \ref{eq:ngp_mix}, we calculate the $\mathcal{N}$-GP prior from this mixture, which is a joint Gaussian mixture distribution over all $N$ modeled time steps.
Now, for determining a posterior predictive distribution, we first partition the prior into a partition containing the time steps to condition on and the remaining time steps.
The posterior weights, mean vectors and covariance matrices can then be directly calculated (see e.g. \citep{bishop2006pattern,petersen2008matrix} for partitioned Gaussian mixtures).
The probability distribution for individual trajectory points can be extracted through marginalization.

\subsection{Experimental Setup}
For the evaluation, we consider scenes from commonly used datasets: \emph{BIWI Walking Pedestrians} (\citep{pellegrini2009you}, scenes: ETH and Hotel), \emph{Crowds by Example} (\citep{lerner2007crowds}, scenes: Zara1 and Zara2) and the \emph{Stanford Drone Dataset} (\citep{robicquet2016learning}, scenes: Bookstore and Hyang).
Following common practice, the annotation rate of each dataset is adjusted to $2.5$ annotations per second.
Further, the evaluation is conducted on trajectories of fixed length $N = N_\text{in} + N_\text{pred} = 8 + 12$.
We trained the LSTM - MDN combination independently on each dataset to generate $\mathcal{N}$-Curve mixture, which model complete trajectories of length $N$.
For training, $80\%$ of each dataset are used.
The initial predictions are then updated by calculating the posterior distributions conditioning on the input's last point $\mathbf{x}_{N_\text{in}}$ (\emph{posterior A}) and on $\{\mathbf{x}_4,\mathbf{x}_{N_\text{in}}\}$ (\emph{posterior B}).
By increasing the number of points, we expect the prediction to adapt towards a given trajectory sample.
For measuring the performance, we apply the \emph{Average Displacement Error} (ADE, \citep{kothari2021human}) according to the standard evaluation approach, using a maximum likelihood estimate.
As the ADE does not provide an adequate measure for assessing the quality of (multi-modal) probabilistic predictions, we use the \emph{Negative Log-Likelihood} (NLL) in addition to the ADE.
This is a common choice for this purpose \citep{bhattacharyya2018accurate,ivanovic2019trajectron}.

\subsection{Results} 
\begin{figure*}[htbp]  
	\centering
	\includegraphics[width=0.3\textwidth]{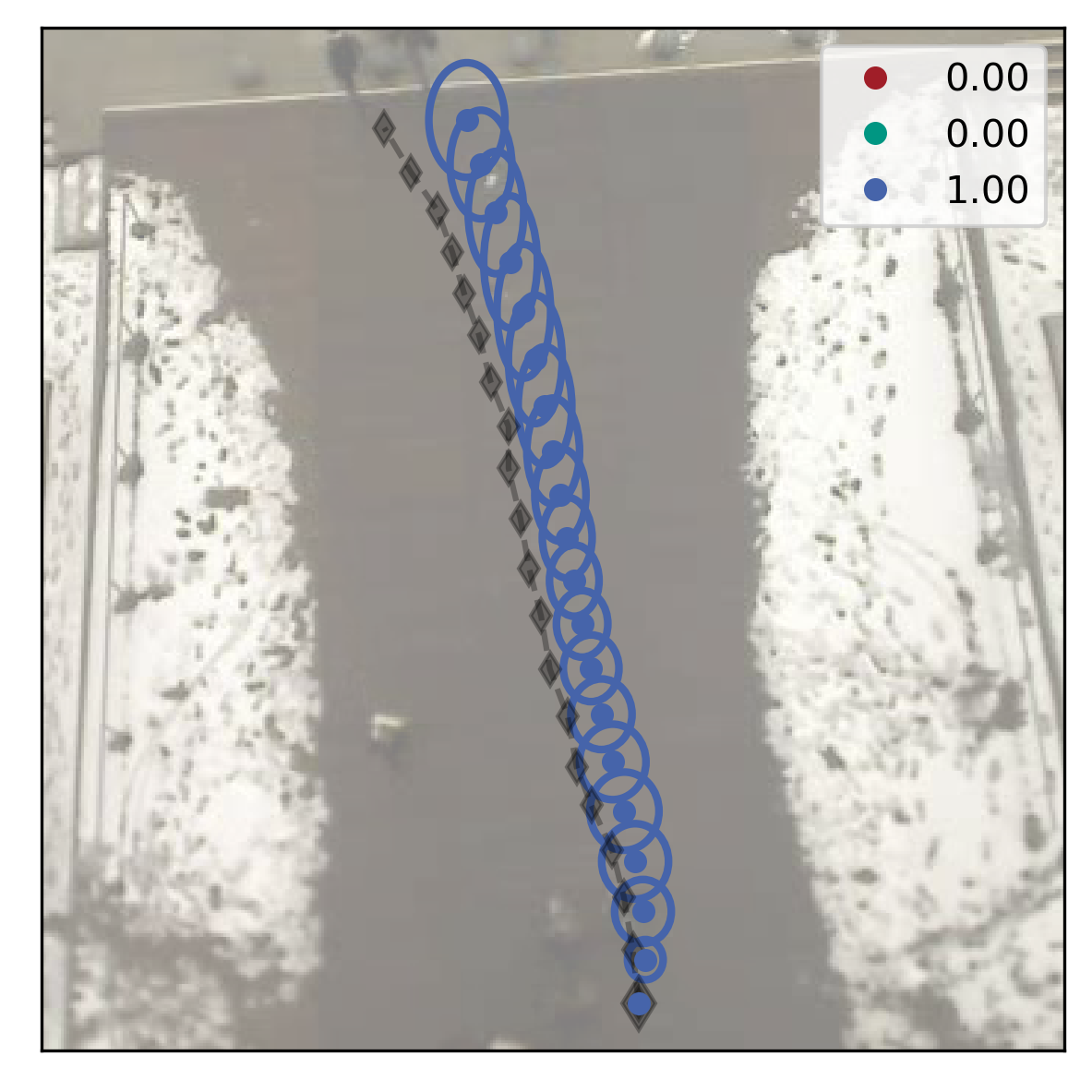}  
	\hspace{9pt}
	\includegraphics[width=0.3\textwidth]{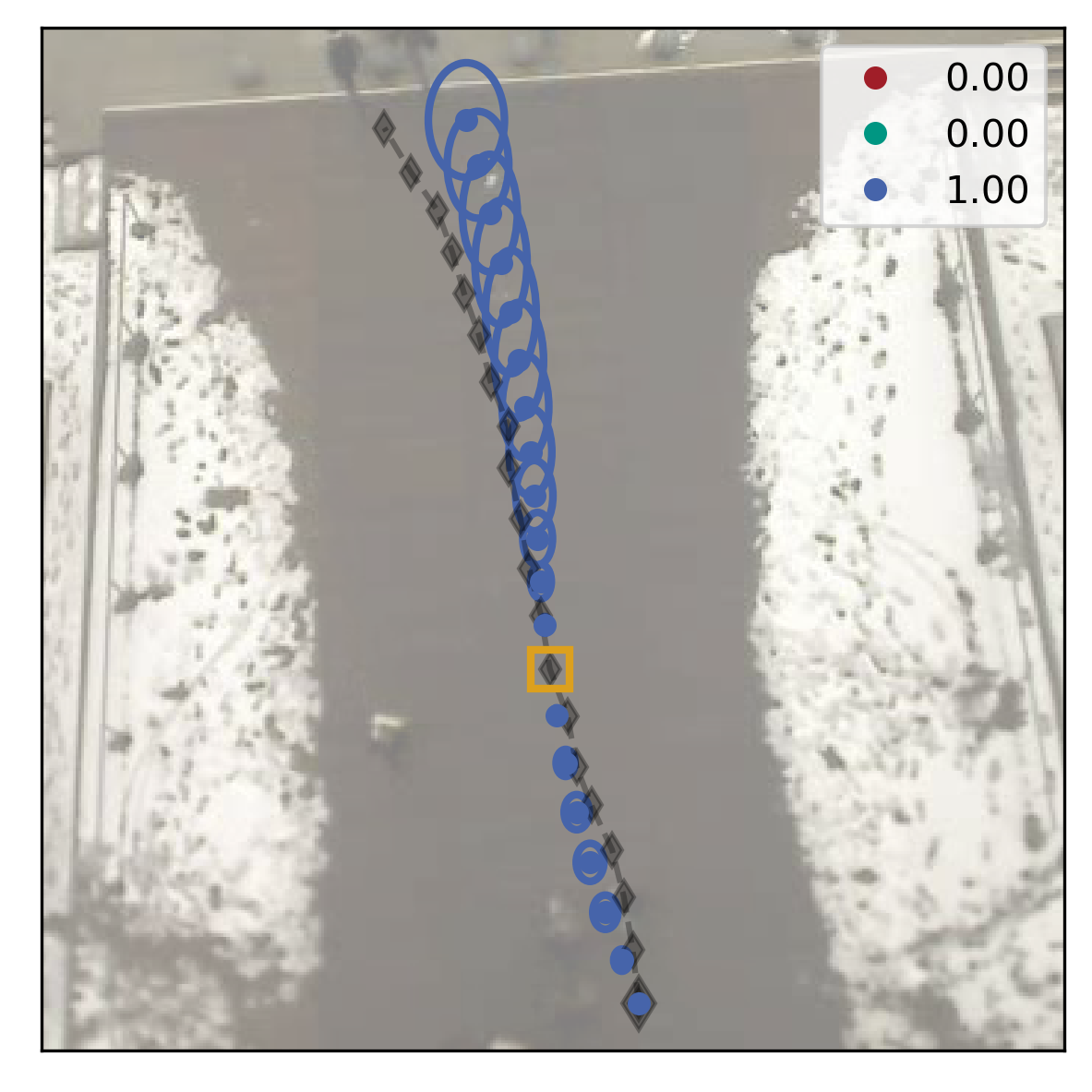}
	\hspace{9pt}
	\includegraphics[width=0.3\textwidth]{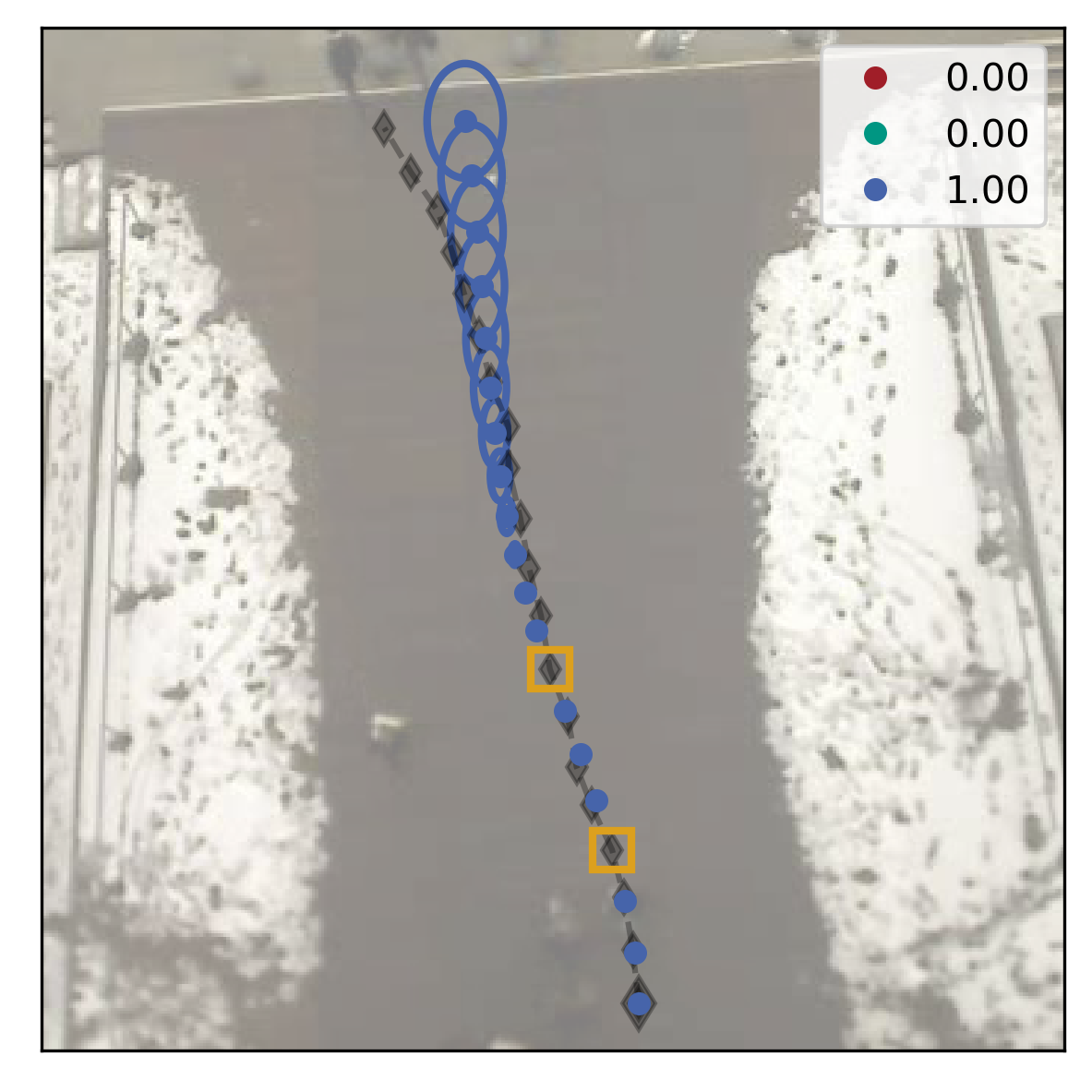}

	\includegraphics[width=0.3\textwidth]{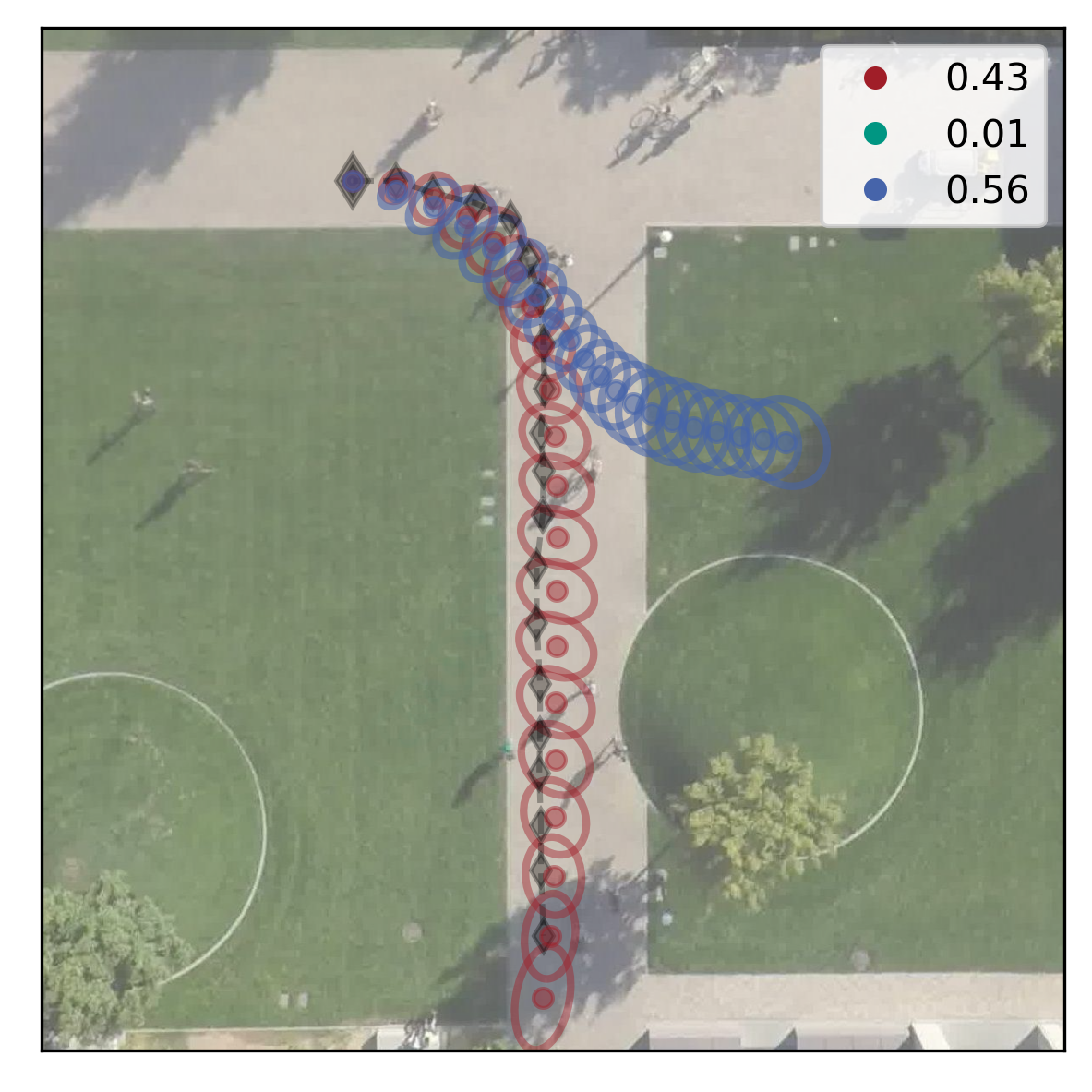}
	\hspace{9pt}
	\includegraphics[width=0.3\textwidth]{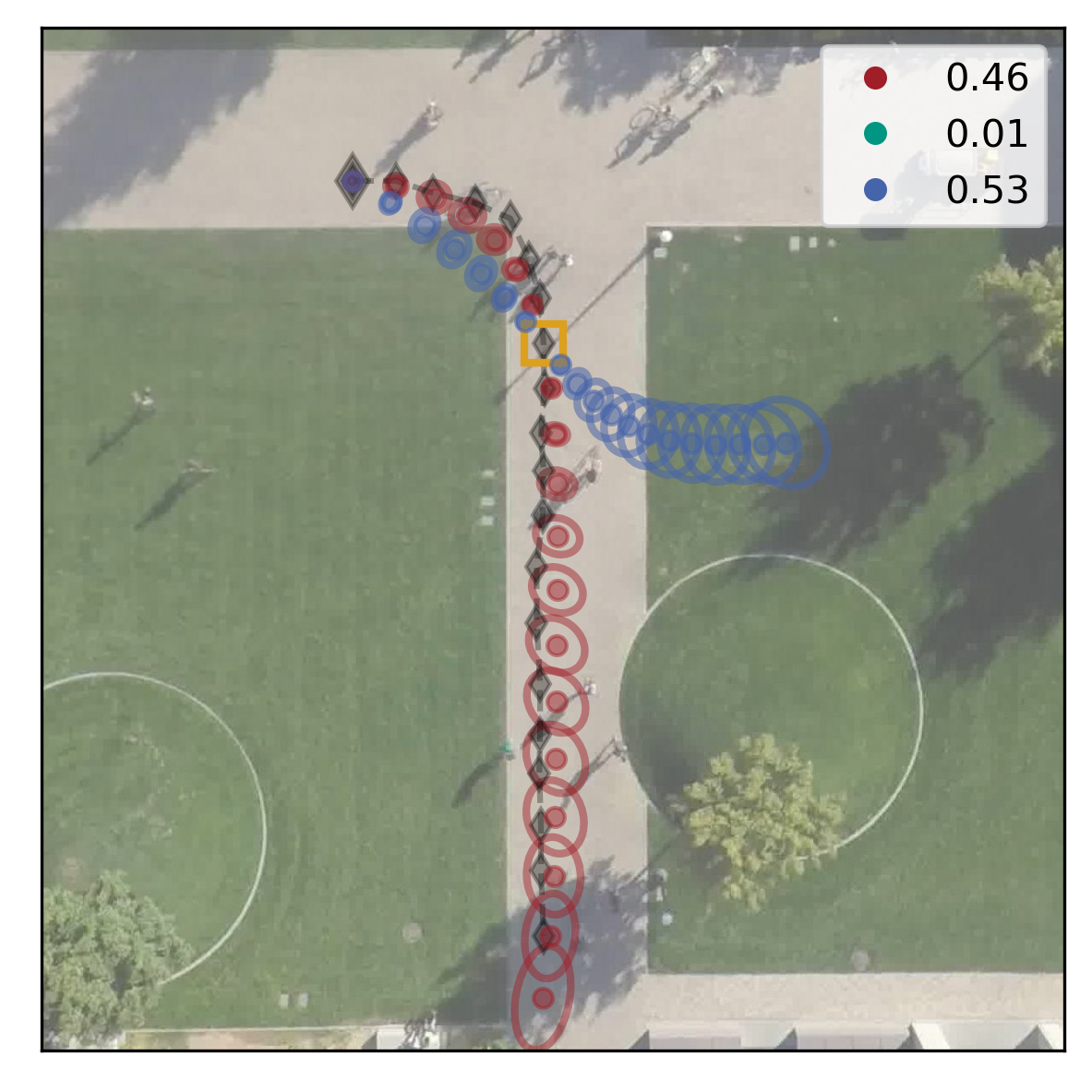}
	\hspace{9pt}
	\includegraphics[width=0.3\textwidth]{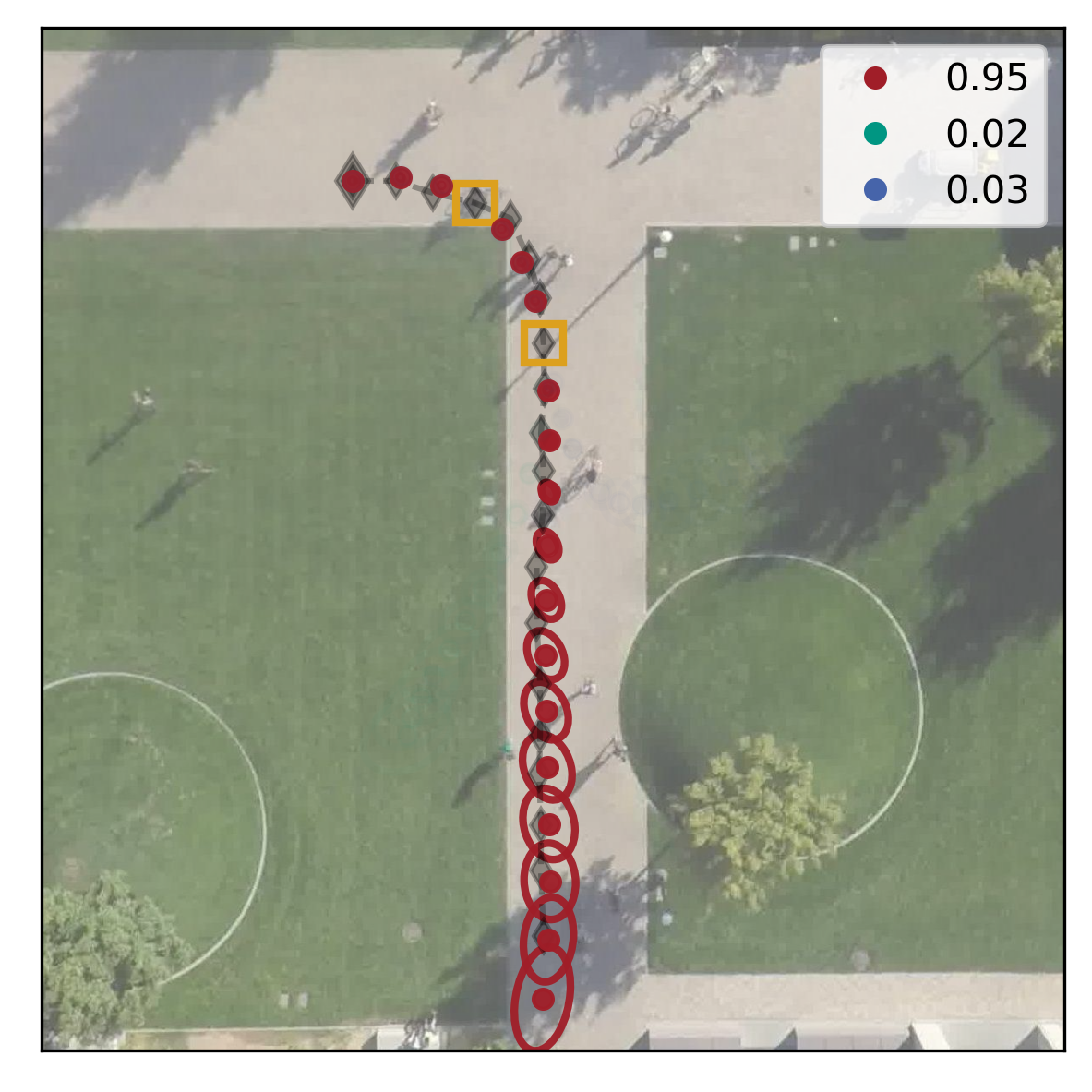}

	\caption{Exemplary cases for improved trajectory prediction through conditioning on $1$ or $2$ points, respectively. Left to right: Prior, posterior A and posterior B. Condition points are indicated by a yellow square. The full ground truth trajectory is depicted in semi-transparent black.}
	\label{fig:good_ex}
\end{figure*}

The quantitative results for the $\mathcal{N}$-GP -- based prediction refinement with respect to the selected performance measures are depicted in Table \ref{tab:quant_res}.
Overall, an increase in performance can be observed when refining the estimate generated by the $\mathcal{N}$-MDN using $1$ and $2$ points, respectively.
This supports our expectation of an increase in prediction performance through adding GP-based Bayesian inference to the $\mathcal{N}$-MDN.
\begin{table*}[htbp] 
	\centering
	\renewcommand{\arraystretch}{1}
	\begin{tabular*}{\textwidth}{@{\extracolsep{\fill}}llccc}
		&& Prior & Posterior A & Posterior B \\
		\hline
		\multirow{2}{*}{ETH}		& ML-ADE & 3.85 / 11.25 & 3.95 / 10.12 & 2.39 / 10.18 \\
									& NLL & 6.51 / 7.58 & 5.43 / 7.08 & -115.09 / 1.70 \\
		\hdashline
		\multirow{2}{*}{Hotel} 	 	& ML-ADE & 5.69 / 17.96 & 4.19 / 17.07 & 2.70 / 16.73 \\
									& NLL & 6.99 / 8.20 & 5.56 / 7.71 & 10.63 / 8.59 \\
		\hdashline
		\multirow{2}{*}{Zara1}  	& ML-ADE & 4.09 / 19.10 & 2.89 / 17.52 & 1.63 / 17.64 \\
									& NLL & 6.83 / 8.18 & 5.27 / 7.63 & -51.93 / 8.15 \\
		\hdashline
		\multirow{2}{*}{Zara2}  	& ML-ADE & 2.98 / 21.38 & 2.64 / 20.07 & 1.69 / 20.05 \\
									& NLL & 6.59 / 8.09 & 5.08 / 7.59 & -60.95 / -1.76 \\
		\hdashline
		\multirow{2}{*}{Bookstore}	& ML-ADE & 4.04 / 17.21 & 3.65 / 15.97 & 2.16 / 16.29 \\
									& NLL & 7.46 / 8.37 & 5.88 / 7.76 & -11.89 / 7.63 \\
		\hdashline
		\multirow{2}{*}{Hyang} 		& ML-ADE & 5.51 / 36.46 & 5.01 / 34.05 & 3.16 / 32.18 \\
									& NLL & 8.21 / 9.42 & 6.65 / 8.86 & -49.30 / 9.05 \\
		\hline
	\end{tabular*}
	\caption{Quantitative results of the (prior) prediction as generated by an $\mathcal{N}$-MDN and posterior refinements. 
	Table entries report the estimation error with respect to the input trajectory (first value) and the trajectory to be predicted (second value), respectively.
	ADE errors are reported in pixels. Lower is better for both performance measures.}
	\label{tab:quant_res}
\end{table*}
Two examples highlighting common cases for a positive effect of the refinement on the prediction performance is given in Fig. \ref{fig:good_ex}.
On the one hand, the refinement can lead to the estimate being pulled closer to the ground truth in the input portion, which expands far into the future prediction.
On the other hand, the refinement can lead to the suppression of inadequate mixture components, which have had high weights assigned to them in the prior distribution.

\begin{figure*}[h]  
	\centering
	\includegraphics[width=0.25\textwidth]{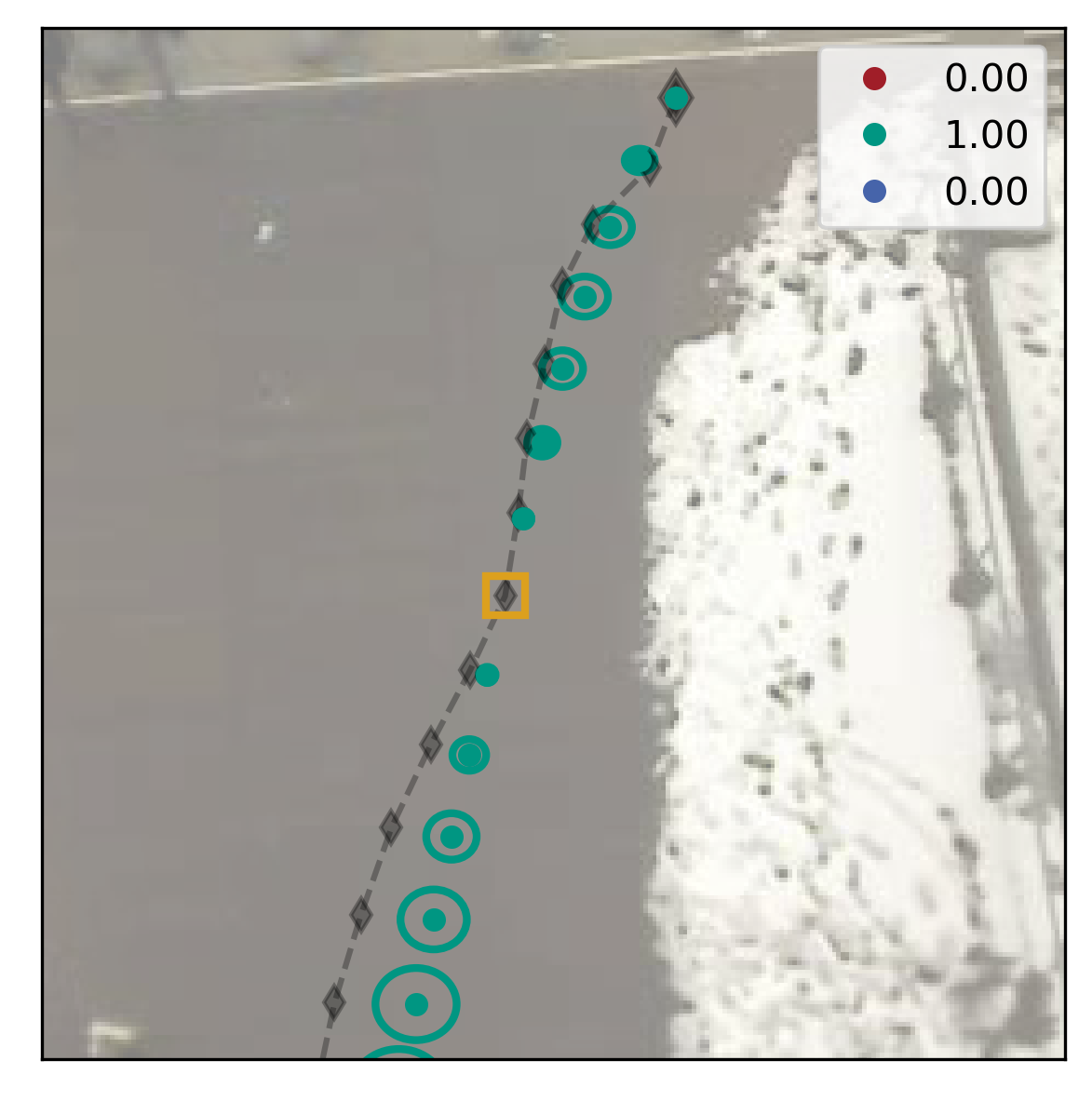}
	\hspace{-10pt}
	\includegraphics[width=0.25\textwidth]{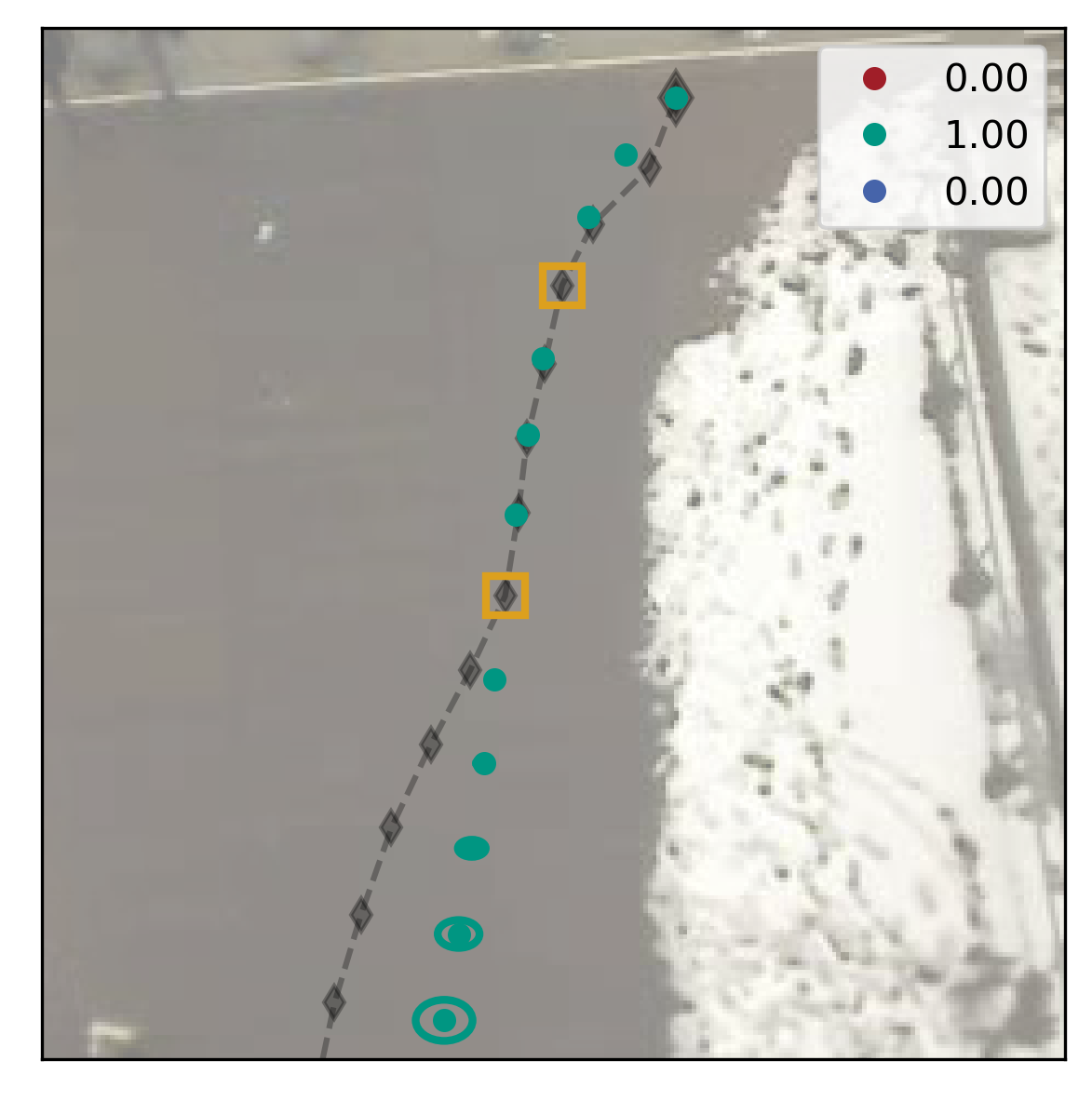}
	\includegraphics[width=0.25\textwidth]{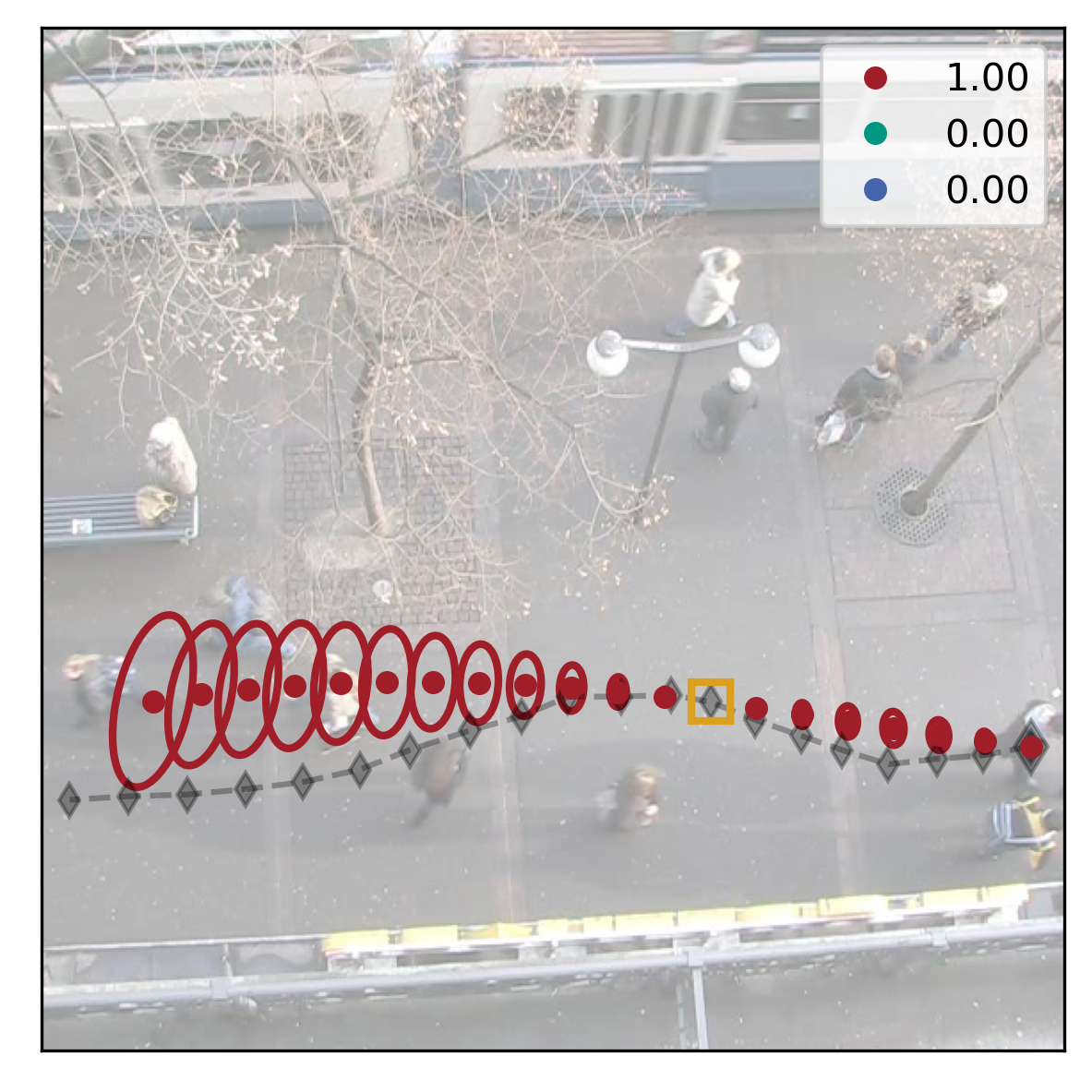}
	\hspace{-10pt}
	\includegraphics[width=0.25\textwidth]{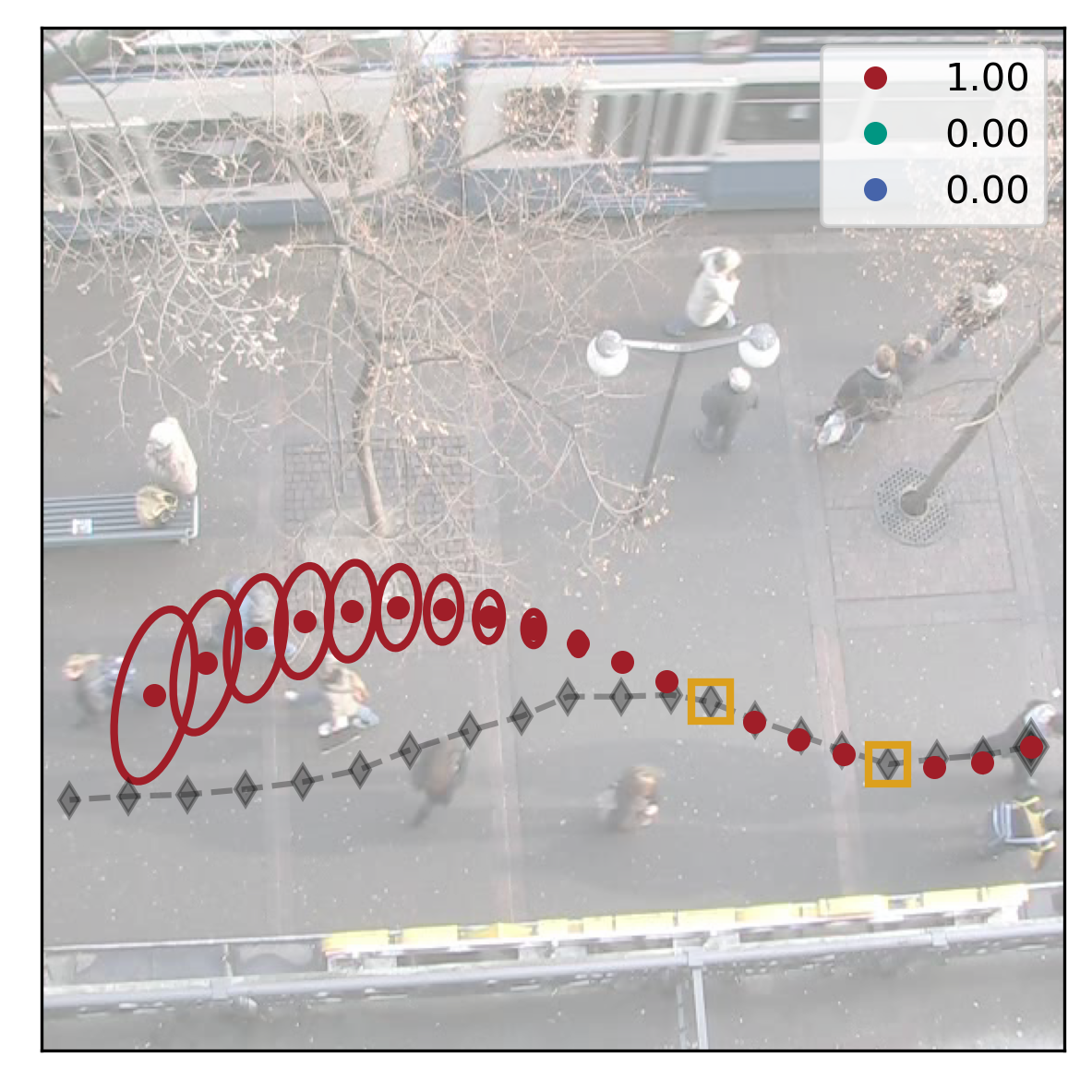}

	\caption{Common cases for refinement leading to a degrade in prediction performance according to the NLL (1 \& 2) and ADE (3 \& 4). 1 \& 3 depict posterior A and 2 \& 4 posterior B. Condition points are indicated by a yellow square. The full ground truth trajectory is depicted in semi-transparent black.}
	\label{fig:bad_ex}
\end{figure*}
\begin{figure*}[h]
	\centering
	\includegraphics[width=0.3\textwidth]{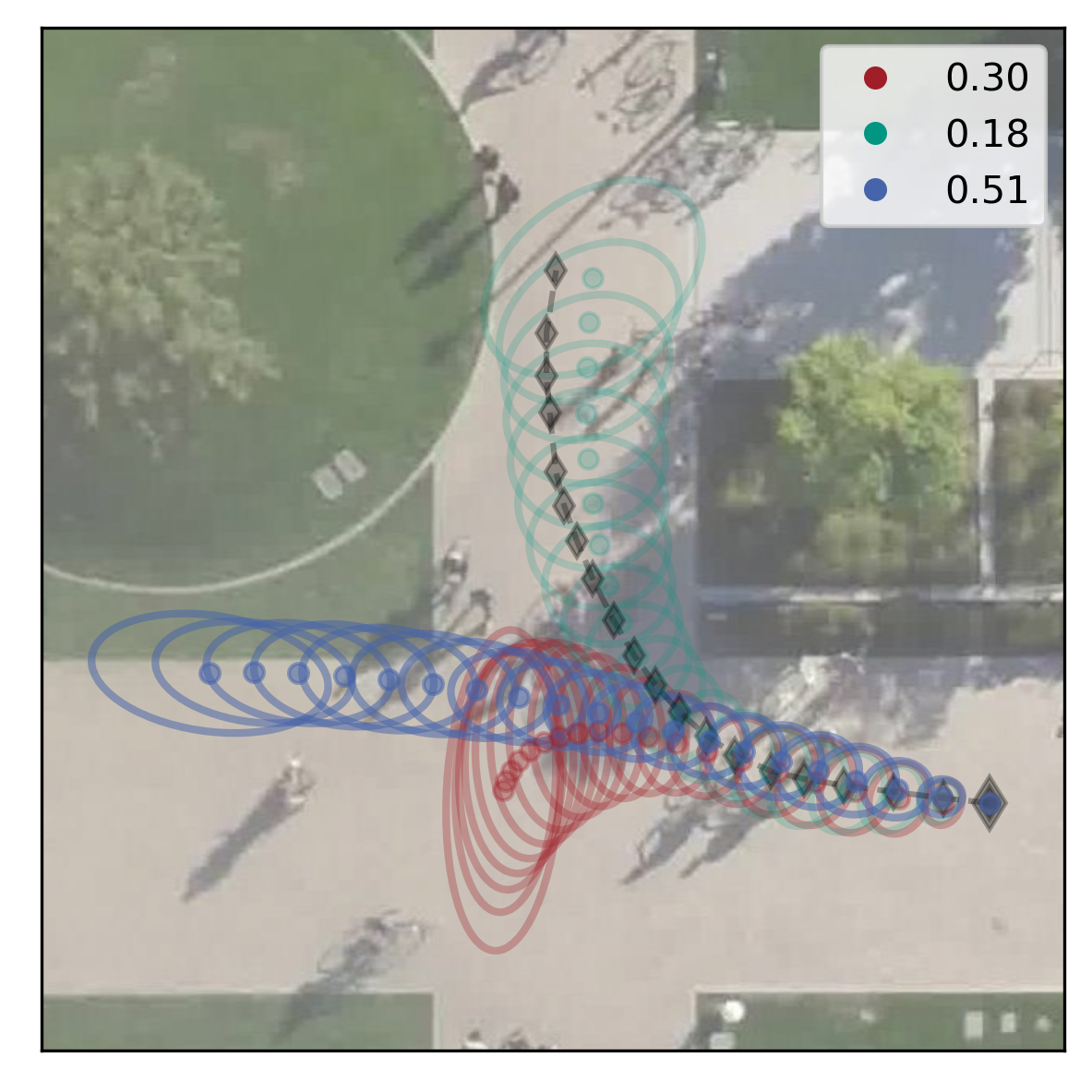}  
	\hspace{9pt}
	\includegraphics[width=0.3\textwidth]{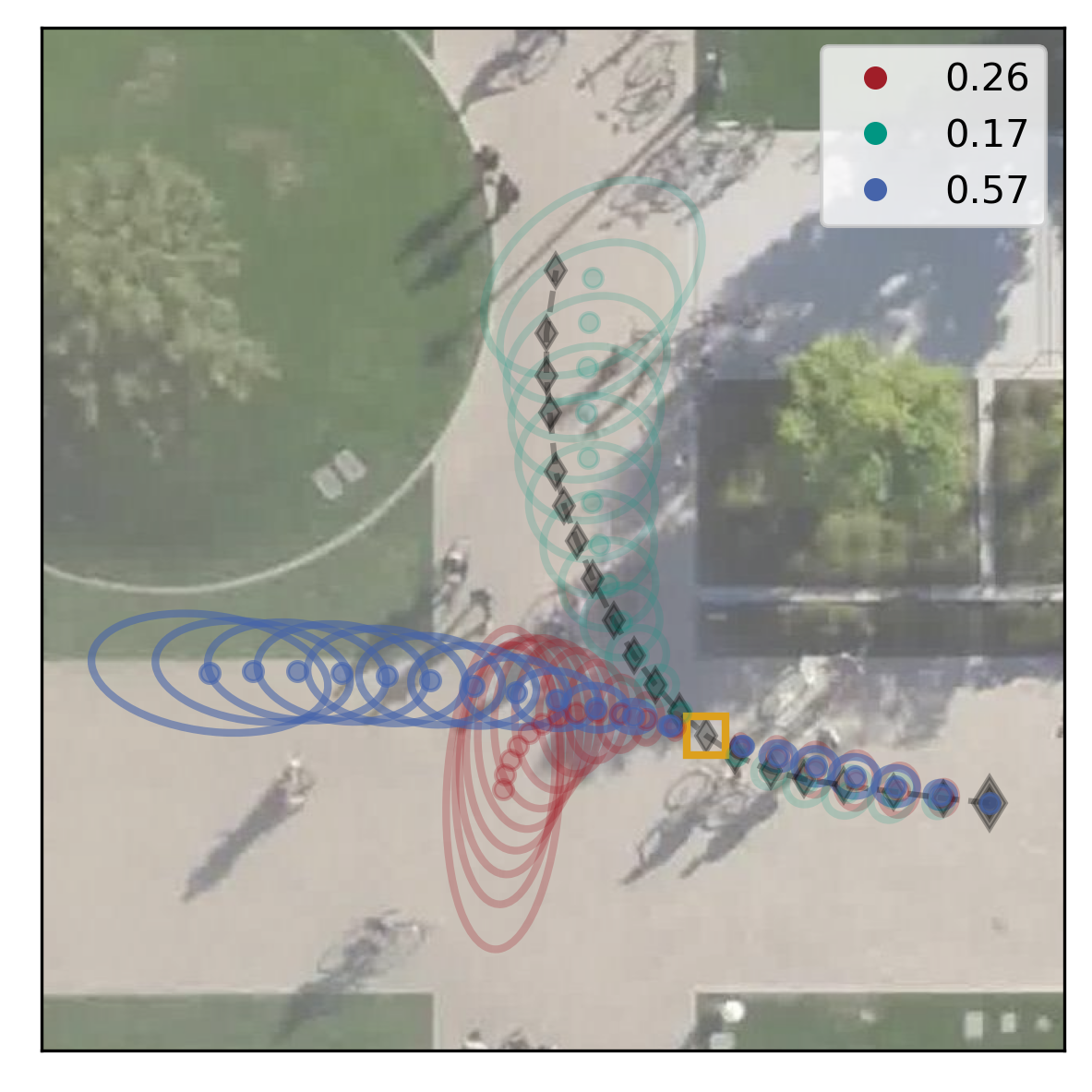}
	\hspace{9pt}
	\includegraphics[width=0.3\textwidth]{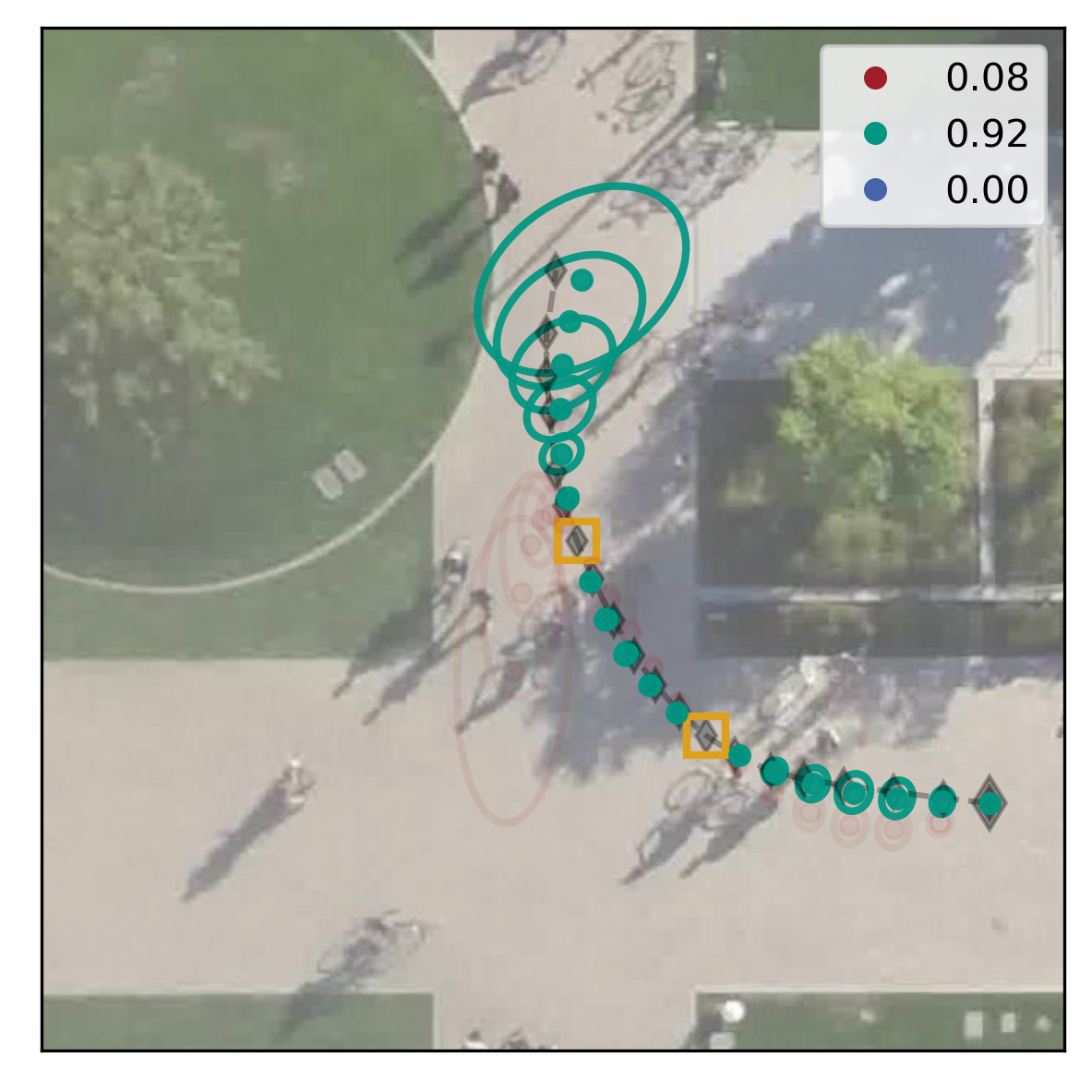}
	\caption{Example for updating the prediction generated by an $\mathcal{N}$-MDN (left) using the last observed trajectory point (center) and an additional point within the prediction time horizon (right).
	Condition points are indicated by a yellow square.
	The full ground truth trajectory is depicted in semi-transparent black.}
	\label{fig:post_cond_ex}
\end{figure*}

Besides the overall performance, it can be seen that in some instances conditioning on $2$ points (posterior B) degrades the performance in comparison to using a single point (posterior A). 
With respect to the NLL, this can be attributed to an increased number of trajectory point variances decreasing or even collapsing.
Then, even minor inaccuracies in the mean prediction result in higher NLL values, even if the estimate is closer to the ground truth.
Looking at the ADE, the loss in performance can most likely be attributed to the enforced interpolation of the condition points, which sometimes leads to unwanted deformations of the mean prediction.
One of the main causes for this is given by the input trajectories commonly being subject to noise.
Examples for both of these cases are depicted in Fig. \ref{fig:bad_ex}. 
It could be noted, that a common approach for dealing with such problems is given by adding an error term to each observed point \citep{gortler2019visual}.
This, however, introduces additional hyperparameters.

Lastly, we briefly showcase our approach considering the prediction update use-case.
An example for the posterior distribution given an additional observation within the prediction time horizon is depicted in Fig. \ref{fig:post_cond_ex}.
While there are initially multiple relevant mixture components (according to their weights), the additional observation leads to the suppression of wrong modes.
Here, the new observation occurs several time steps after the last original input.
Using the $\mathcal{N}$-GP, the updated prediction can be directly calculated without requiring an additional pass through the $\mathcal{N}$-MDN.

\section{Summary}
In this paper, we presented an approach for enabling full Bayesian inference without the need for Monte Carlo methods on top of the $\mathcal{N}$-Curve Mixture Density network ($\mathcal{N}$-MDN), which is a regression-based probabilistic sequence model and outputs (mixtures of) probabilistic B\'ezier curves ($\mathcal{N}$-Curves).
In our approach, the $\mathcal{N}$-MDN is embedded in the GP framework as a generator for prior distributions.
For this, we first showed that $\mathcal{N}$-Curves are a special case of Gaussian processes (denoted as $\mathcal{N}$-GP) and then derived mean and kernel functions for the univariate, multi-variate and multi-modal cases.
In our evaluation on the task of human trajectory prediction, we showed that using the $\mathcal{N}$-GP, predictions generated by an $\mathcal{N}$-MDN can be improved by conditioning on different subsets of the original input.
Additionally, we looked briefly into practical applications of the approach, focusing on updating predictions generated by the $\mathcal{N}$-MDN in light of new observations within the prediction time horizon.
Using our approach, such updates do not require any additional passes through the $\mathcal{N}$-MDN, due to the use of the GP framework.
Further, missing intermediate observations are inherently handled by the $\mathcal{N}$-GP.

\bibliography{bibliography}

\end{document}